%% file: main.tex
\documentclass[10pt,twocolumn,letterpaper]{article}

\usepackage{iccv}              
\usepackage{multirow}
\usepackage[normalem]{ulem}
\useunder{\uline}{\ul}{}
\usepackage{booktabs}
\usepackage{pifont}

\usepackage{marvosym}


\usepackage[accsupp]{axessibility}  

\definecolor{iccvblue}{rgb}{0.21,0.49,0.74}
\usepackage[pagebackref,breaklinks,colorlinks,allcolors=iccvblue]{hyperref}

\definecolor{myPurple}{rgb}{0.4, .0, .8}
\definecolor{myGreen}{rgb}{0, .8, .3}
\definecolor{myRed}{rgb}{0.8, .2, .2}
\definecolor{myOrange}{rgb}{0.8, 0.45, 0.0}
\definecolor{myBlue}{rgb}{.0, .0, 1.0}

\def\paperID{12218} 
\def\confName{ICCV}
\def\confYear{2025}

\newcommand{\modelname}{{TeRA}}

\title{\modelname: Rethinking Text-guided Realistic 3D Avatar Generation}

\author{Yanwen Wang$^{1*}$, Yiyu Zhuang$^{1*}$, Jiawei Zhang$^{1}$, Li Wang$^{1}$, Yifei Zeng$^1$, \\ Xun Cao$^1$, Xinxin Zuo$^2$, Hao Zhu$^{1}$$\textsuperscript{\Letter}$\\
$^{1}$Nanjing University  \quad $^2$Concordia University
\\
}

\begin{document}

\twocolumn[{%
\renewcommand\twocolumn[1][]{#1}%
\maketitle
\vspace{-0.35in}
\begin{center}
    \centering
    \includegraphics[width=0.95\linewidth]{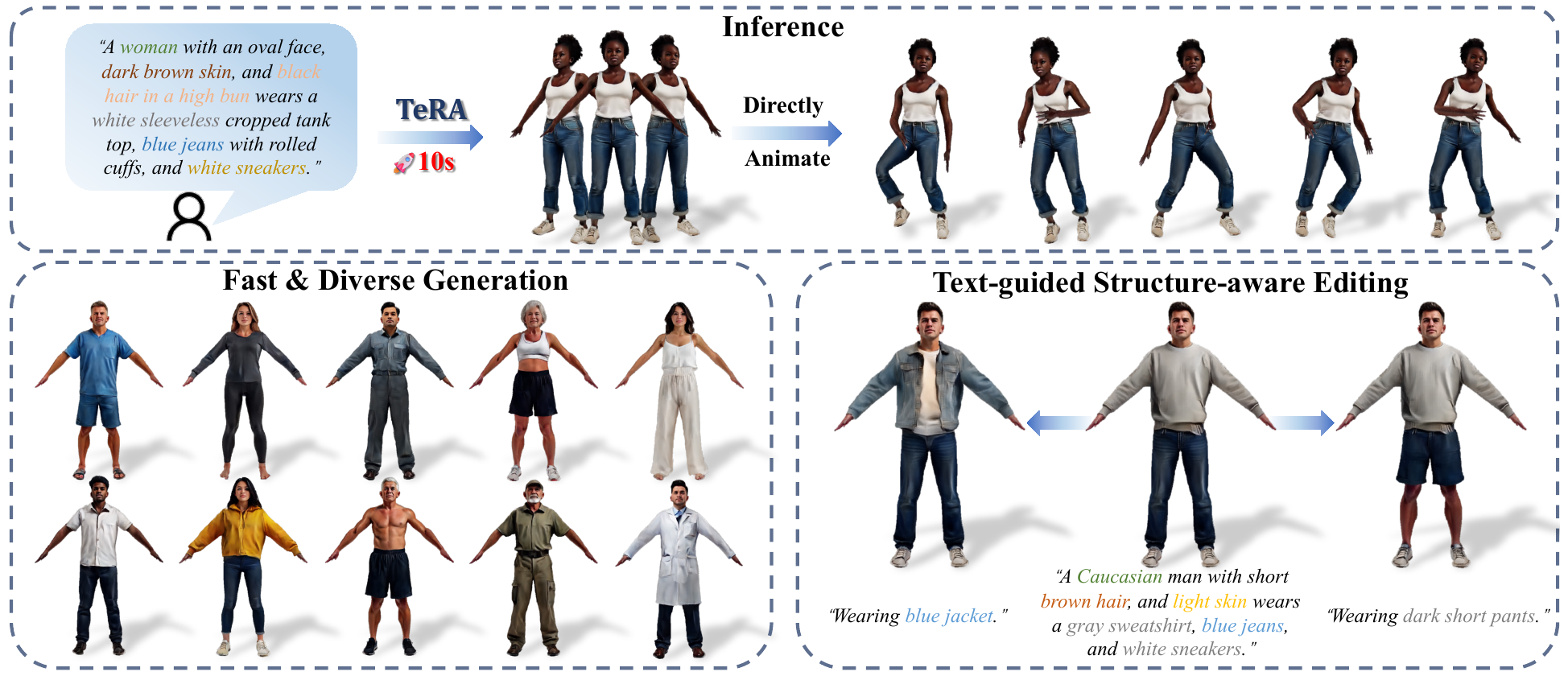}
    \vspace{-0.05in}
    \captionof{figure}{We propose TeRA, the first latent diffusion model specifically designed for text-guided 3D avatar generation. 
    TeRA achieves superior inference speed, text-to-3D alignment, and visual quality, while naturally supporting text-guided structure-aware editing.
    }
\label{fig:title}
\end{center}
}]

\renewcommand{\thefootnote}{}
\footnotetext{* These authors contribute equally to this work.}

\input{sec/0_abs}
\input{sec/1_intro}
\input{sec/2_related}
\input{sec/3_method}
\input{sec/4_exp}
\input{sec/5_con}

\subsection*{Acknowledgements}
This work was supported in part by the National Key Research and Development Program of China under Grant 2022YFF0902401, in part
by the National Natural Science Foundation of China under Grant 62025108.

{
    \small
    \bibliographystyle{ieeenat_fullname}
    \bibliography{main}
}

\input{sec/X_suppl}    


\end{document}

%% file: sec/0_abs.tex
\begin{abstract}
Efficient 3D avatar creation is a significant demand in the metaverse, film/game, AR/VR, etc.  
In this paper, we rethink text-to-avatar generative models by proposing \modelname, a more efficient and effective framework than the previous SDS-based models and general large 3D generative models.
Our approach employs a two-stage training strategy for learning a native 3D avatar generative model. Initially, we distill a decoder to derive a structured latent space from a large human reconstruction model. Subsequently, a text-controlled latent diffusion model is trained to generate photorealistic 3D human avatars within this latent space.  
TeRA enhances the model performance by eliminating slow iterative optimization and enables text-based partial customization through a structured 3D human representation. 
Experiments have proven our approach's superiority over previous text-to-avatar generative models in subjective and objective evaluation. The project page is available at \url{https://yanwen-w.github.io/TeRA-Page/}.
\end{abstract}


%% file: sec/1_intro.tex
\section{Introduction}
\label{sec:intro}

With the explosive growth of the metaverse, film/game production, and the AR/VR industry in recent years, the creation of rapid, convenient, and efficient 3D human avatars has emerged as a critical bottleneck.  Conventional approaches to 3D human avatar reconstruction often involve intricate and time-consuming modeling techniques using a single camera~\cite{xiu2023econ, zheng2021pamir, zhu2021detailed, saito2019pifu}, camera arrays~\cite{guo2019relightables, debevec2012light, alexander2010digital}, or range sensors~\cite{yu2021function4d, yu2018doublefusion}.  On another line of research, large 3D generative models \cite{clay, direct3d, xiang2025structured} have recently emerged as an efficient method for producing 3D models from image or text descriptions.  Nevertheless, experimental results indicate that all existing 3D generative large models fail to produce plausible results for \textit{photorealistic} 3D human avatars.  Such defect is attributed to a profound style bias in the training samples, where there is an overabundance of designed cartoon-like character models and a dearth of high-precision realistic human models. 

For the creation of \textit{photorealistic} 3D human avatars, the state-of-the-art approach leverages the score distillation sampling (SDS) strategy~\cite{poole2022dreamfusion}. The core idea of SDS lies in its utilization of pre-trained 2D diffusion models to steer the optimization process for generating 3D models, without the necessity for any 3D data for training. The 3D priors employed in SDS-based methods are derived from pre-trained 2D vision-language large models~\cite{poole2022dreamfusion}, rich in photorealistic human features. Nevertheless, the inherent absence of explicit 3D structures in 2D diffusion models poses a challenge in ensuring multi-view consistency. This leads to the suboptimal quality of 3D human avatars generated within the SDS framework. Furthermore, the iterative distillation procedure inherent in SDS methods often necessitates a substantial amount of time to complete the optimization process.  These limitations of the SDS route prevent it from achieving efficient and robust 3D avatar generation. 

In this paper, we introduce TeRA, a feedforward text-to-avatar generative model tailored for the efficient, realistic, and editable creation of 3D humans driven by text. To tackle the challenge of insufficient 3D human data, we leverage a collaborative approach that integrates large vision-language and language models, providing highly accurate and detailed appearance descriptions for HuGe100K~\cite{zhuang2025idol}, an extensive 3D human dataset.  Regarding network architecture and training, we utilize a two-stage feedforward prediction model. The first stage employs an autoencoder to extract a structured and readily generable latent space from a comprehensive human reconstruction model. Subsequently, the second stage trains a text-controlled latent diffusion model within this latent space, generating diverse and lifelike 3D human models.  We noticed that directly connecting the diffusion model to the encoder results in significantly diminished performance, primarily due to posterior collapse. To address this issue, we propose a distillation module that improves the generated quality and reduces the training resources.  Furthermore, by incorporating a structured human representation, we have enabled fine-grained editing of a partially customizable 3D avatar through text descriptions.


Our framework markedly improves the performance of the text-to-avatar model. Firstly, by embracing a single-pass prediction framework, TeRA obviates the necessity for the slow and cumbersome iterative optimization processes characteristic of SDS-based methods. Secondly, incorporating a structured 3D human representation enables text-based partial customization, significantly enhancing usability. Lastly, our model exhibits exceptional generation quality and text-model alignment, outperforming SDS-based approaches and general 3D generative large models. Comprehensive user studies and qualitative/quantitative experiments substantiate our TeRA's superior performance.

The contributions of this paper can be summarized as:

\begin{itemize}
    \item We propose the pioneering text-to-3D avatar generative model built upon the latent diffusion model framework.  In terms of speed, text-model alignment, and rendering quality, it surpasses previous state-of-the-art models that leverage scored distillation sampling.
    \item A distillation module that links the diffusion model to the VAE encoder has been introduced, serving as an essential component for generating high-quality avatar models.
    \item By introducing a structured 3D human representation, structure-aware editing is achieved for a partially customizable 3D avatar.
\end{itemize}




%% file: sec/2_related.tex
\section{Related Work}
\label{sec:related}

\subsection{2D Diffusion-based Generative Model}
Recent years have witnessed remarkable progress in vision-language technologies, driven by breakthroughs in cross-modal representation learning~\citep{radford2021learning} and generative models~\citep{ho2020ddpm, rombach2022ldm, latentdiffusion, lipman2023flow, zhu2024champ}. These approaches, trained on massive-scale text-image datasets, demonstrate unprecedented capability in understanding and synthesizing visual content. Such advancements have propelled significant improvements in text-to-image generation systems~\citep{saharia2022photorealistic, ramesh2022dalle2, balaji2022ediffi, rombach2022high} and laid the foundation for text-to-video synthesis~\citep{blattmann2023stable, liu2024sora, guo2024i2v, ma2024latte}.
With the large-scale data containing bilions of image-text pairs and video-text pairs, the diffusion model shows great understanding of general objects and enabling the synthesis of high-quality and diverse objects.
Furthermore, many works have exploring the controlable generalization with addition condition, including camera motion \cite{zheng2024cami2v, hou2024training, cheong2024boosting, he2024cameractrl, jin2025flovd} or others \cite{zhao2023uni, zhang2023adding, asmussen1997controlled, bansal2023universal}.

\subsection{Text-to-3D Generation} 
Recent text-to-3D generation methods can be broadly categorized into two main approaches: feedforward generation and optimization-based generation.
\textbf{Feedforward generation} methods employ a variety of 3D representations, including point clouds~\cite{achlioptas2018learning, shu20193d}, voxel grids~\cite{wu2016learning}, meshes~\cite{cheng2019meshgan}, implicit radiance fields~\cite{chan2022efficient, an2023panohead, zhuang2022mofanerf, he2024head360}, and 3D Gaussian Splatting (3DGS)\cite{kerbl3Dgaussians, kirschstein2024gghead, zhuang2024towards}. GAN-based approaches leverage conditional GANs\cite{chen2019text2shape, liu2022towards, tian2023shapescaffolder, wei2023taps3d} to generate 3D assets, but they often struggle with limited diversity and suboptimal quality. Recently, diffusion-based methods for native 3D generation~\cite{clay, direct3d, xiang2025structured, TripoSR2024} have shown promise by directly generating 3D shapes and textures from text prompts. However, these methods require high-quality, large-scale 3D asset datasets to achieve satisfactory results.
\textbf{Optimization-based generation} methods adopt a per-prompt generation strategy by distilling 3D knowledge from rich priors learned in the 2D domain. For example, early approaches utilized CLIP guidance~\cite{radford2021learning} to generate the multi-view information~\cite{jain2022zero, wang2022clip, mohammad2022clip, michel2022text2mesh}. More recent methods employ score distillation sampling (SDS) \cite{poole2022dreamfusion} to transfer the high-quality rendering capabilities of state-of-the-art text-to-image models~\cite{poole2022dreamfusion, wang2023score, metzer2023latent, wang2023prolificdreamer, chen2023fantasia3d, tang2023dreamgaussian, chen2024text, yi2023gaussiandreamer}. Despite their advancements, these methods are hindered by prolonged per-scene optimization times and often produce cartoonish or multi-face 3D outputs.

\subsection{Text-to-3D Avatar Generation}
By incorporating human prior such as SMPL~\cite{SMPL:2015} and SMPL-X~\cite{pavlakos2019expressive}, the 3D human generation literature has emerged as a distinct subfield within text-to-3D research~\cite{hong2022eva3d, cao2024dreamavatar, jiang2023avatarcraft, zeng2023avatarbooth, zhang2023text, gao2023textdeformer, zhao2023zero, wang2024disentangled, zhuang2024dagsm}.
AvatarCLIP~\cite{hong2022avatarclip} combines CLIP guidance with SMPL templates to generate 3D avatars.
DreamWaltz~\cite{huang2023dreamwaltz} introduces 3D-aware skeleton conditioning and occlusion-aware SDS to mitigate the Janus (multi-face) problem.
DreamHuman~\cite{kolotouros2023dreamhuman} utilizes imGHUM~\cite{imGHUM} to encode pose- and shape-conditioned signed distance fields, enhancing neutral human reconstruction.
HumanNorm~\cite{huang2023humannorm} enhances geometric details by fine-tuning a text-to-depth/normal diffusion models to provide explicit structural constraints. 
AvatarVerse~\cite{zhang2023avatarverse} fine-tunes the ControlNet~\cite{zhang2023adding} branch with DensePose~\cite{guler2018densepose} as an SDS source for multi-view generation. 
HumanGaussian~\cite{liu2024humangaussian} integrates skeleton and depth maps to regulate the 3DGS embedded on the SMPL-X template, enabling efficient rendering.
TADA~\cite{liao2024tada} applies displacement maps to the SMPL-X shape and texture UV map to represent 3D avatars, optimizing them through a hierarchical rendering approach with SDS.
However, these optimization-based methods often suffer from significant drawbacks, including long optimization times—sometimes requiring several hours per scene—and the generation of unrealistic results, such as cartoon-like appearances and oversaturation.
Building on the insights from native 3D generation models, we propose a feedforward generation pipeline for 3D human avatars. This approach significantly improves both the generated results' efficiency and realism.

%% file: sec/3_method.tex
\section{Method}
\label{sec:method}
\begin{figure*}[t]
    \centering
    \includegraphics[width=1.0\textwidth]{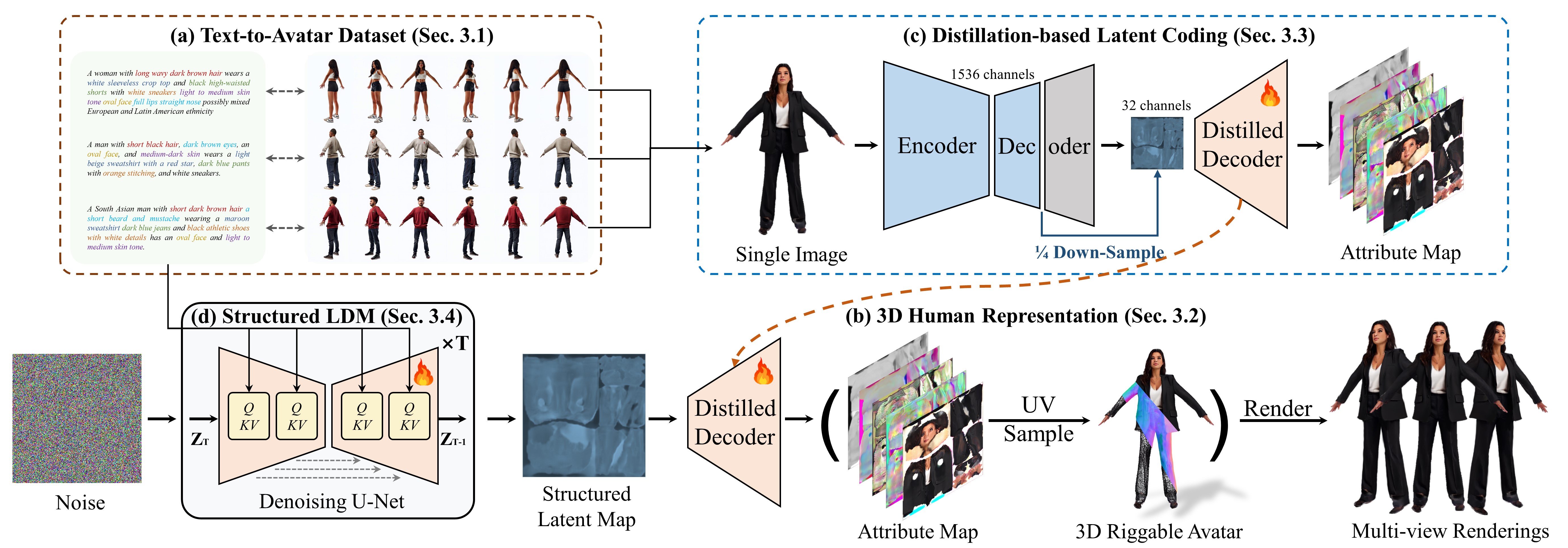}
    \caption{Overall method. (a) Given the annotated multi-view human dataset, we train a text conditioned 3D avatar generative model. (b) The model is established upon a structured 3D human representation. The model training includes two stages: (c) firstly, a decoder is required by distilling a pretrained 3D human reconstruction model; (d) secondly, a structured latent diffusion model (LDM) is trained to generate structured latent maps from noises.
    }
    \label{fig:pipe}
    \vspace{-0.1in}
\end{figure*}

This section introduces TeRA, an efficient, realistic, and editable text-guided 3D human generation model. The text-to-avatar data creation is outlined in Sec.~\ref{sec: data}, followed by 3D avatar representation in Sec.\ref{sec: rep}. Subsequently, we elaborate on the network architecture and training strategy, encompassing a two-stage latent compression approach detailed in Sec.~\ref{sec: decoder}, and a structured latent diffusion model presented in Sec.~\ref{sec: diffusion}. Lastly, Sec.~\ref{sec: edit} discusses structured-aware editing, which allows for fine-grained customization of a generated 3D human avatar.

\subsection{Text-to-Avatar Dataset}
\label{sec: data}

To train a 3D Avatar generation model, the primary challenge lies in establishing a large-scale and diverse text-to-avatar dataset. The HuGe100K~\cite{zhuang2025idol} 
dataset, comprising 100k photorealistic multi-view 3D human models, effectively fulfills the requirements for extensive and varied data. However, it suffers from a lack of text annotations. To address this, we enhance the HuGe100K dataset by incorporating semantic annotations, creating a comprehensive large-scale text-to-avatar dataset.

Early annotation of 3D objects often relies on vision-language models such as BLIP~\cite{li2022blip} or CLIP~\cite{radford2021learning}. However, these models struggle to generate detailed and accurately aligned descriptions, which limits the diversity and precision of text-to-3D generative models.
Recently, large vision-language models(VL model)\cite{yang2024qwen2, bai2025qwen2, achiam2023gpt} have demonstrated excellent performance in image understanding tasks. Therefore, following \cite{ge2024visual}, we adopt a collaborative annotation approach using a large vision-language model Qwen2.5-VL\cite{bai2025qwen2}  and a large language model Qwen2.5\cite{yang2024qwen2} to annotate multi-viewpoint human image data.

As illustrated in Fig.~\ref{fig:pipe} (a), we first input front/back/right/left views of a human into the Qwen2.5-VL, obtaining comprehensive raw descriptions of various body parts through carefully designed prompts, including facial features, upper and lower clothing, shoes, and more. 
Subsequently, these raw descriptions are processed by the Qwen2.5 to extract essential information, producing a concise and precise description of no more than 40 words. Finally, Qwen2.5 further refines and condenses the content into five succinct phrase-based descriptions of varying lengths with the longest being no more than 16 words and the shortest containing at least 8 words. Our trained text-guided 3D human generation model demonstrates high textual consistency thanks to the meticulous and accurate text annotations. Additional details regarding the data annotation can be found in \textit{the supplementary material}.

\subsection{3D Human Representation}
\label{sec: rep}

After creating the dataset, our next step is establishing text-to-avatar generative models. The first issue is deciding how to represent 3D human avatars. In this paper, we follow prior works~\cite{zhang20243gen, zhuang2025idol, zhang2025fate} to represent 3D human avatars with UV-structured 3D Gaussians.  We will begin by providing preliminary knowledge about SMPL-X~\cite{pavlakos2019expressive} and 3D Gaussian Splatting (3DGS)~\cite{kerbl3Dgaussians}, followed by an introduction to the concrete representation settings we employ.

\noindent\textbf{SMPL-X.} is a deformable 3D parametric human model with excellent driving performance and decoupled shape and pose control, currently widely applied in human-driven reconstruction and generation tasks. SMPL-X generates a 3D human mesh using shape parameter \(\beta\), pose parameter \(\theta\), and expression parameter \(\psi\). The generated mesh consists of 10,475 vertices and 54 joints. The deformed human mesh \(M(\beta, \theta, \psi)\) is derived from the mesh \(T(\beta, \theta, \psi)\) in the canonical space through linear blend skinning (LBS). The process is formulated as:

\begin{equation}
M(\beta, \theta, \psi) = \text{LBS}(T(\beta, \theta, \psi),J(\beta), \theta, \psi, W)
\end{equation}

\noindent where \(J(\beta)\) represents the positions of the key joints, and \(W\) denotes the skinning weights. The canonical mesh \(T(\beta, \theta, \psi)\) is obtained using the following formula:

\begin{equation}
T(\beta, \theta, \psi) = T_c + B_s(\beta) + B_e(\psi) + B_p(\theta)
\end{equation}

\noindent where \(T_c\) is the template human mesh, 
and \(B_s(\beta)\), \(B_e(\psi)\), \(B_p(\theta)\) represent shape-dependent, expression-independent, and pose-dependent deformations, respectively.

\noindent\textbf{3D Gaussian Splatting.}
3D Gaussian is an explicit representation for 3D scenes, composed of a set of 3D Gaussian primitives that can be real-time rendered via differentiable rendering. Each primitive consists of the following four properties: position \( \mu \), opacity \( \alpha \), color \( c \), and covariance matrix \( \Sigma \). In practice, the covariance matrix is typically assumed to be \( \Sigma = RSS^T R^T \), where \( S \) represents the size of the Gaussian ellipsoid, and \( R \) is its rotation matrix.

By applying a view transformation, the 3D Gaussian primitives are projected onto the imaging plane, resulting in a set of 2D Gaussian ellipses. The final imaging process is as follows:

{\small
\begin{equation}
c(p) = \prod_{i \in N} \left( c_i \sigma_i \prod_{j=1}^{i-1} (1 - \sigma_j) \right),\quad \sigma_i = \alpha_i G(p, \mu_i, s_i, r_i),
\end{equation}
}

\noindent where \( p \) is the query point position, and \( \mu_i \), \( s_i \), \( r_i \), \( c_i \), and \( \alpha_i \) represent the position, scale, rotation, color, and opacity of the \( i \)-th Gaussian, respectively. \( G(p, \mu_i, s_i, r_i) \) represents the value of the \( i \)-th Gaussian at the point \( p \).

\noindent\textbf{Structured Gaussians for 3D Human.} 
We represent the 3D human body using a structured Gaussian attribute map, where each Gaussian's attributes are stored in a UV space aligned with the SMPL-X mesh. 
Initially, the position of each 3D Gaussian \( \hat{\mu}_k \) is set to the densified SMPL-X mesh vertices. The scale \( \hat{s}_k \) is defined by the relative distance to neighboring Gaussians, and the rotation \( \hat{r}_k \) is aligned with the local tangent frame of the 3D surface. 
A neural network then predicts offset values  \( \{ \delta_{\mu_k}, \delta_{r_k}, \delta_{s_k} \} \) for position, rotation, and scale, as well as the color \( c_k \) and opacity \( \alpha_k \) of each Gaussian. The final attributes of the Gaussians are computed as:
\begin{align}
    \mu_k &= \hat{\mu}_k + \delta \mu_k \\
    r_k &= \hat{r}_k \cdot \delta_{r_k} \\
    s_k &= \hat{s}_k \cdot \delta_{s_k}
\end{align}  

The complete set of attributes, including \( \mu_k \), \( r_k \), \( s_k \), \( c_k \), and \( \alpha_k \), are stored in a multi-channel attribute map within the UV space of the SMPL-X mesh. As shown in Fig.~\ref{fig:pipe} (b), this attribute map is transformed into 3D Gaussian Human through UV sampling and then rendered into images. 
In our approach, the attribute map serves as the output of the generative network, enabling flexible editing and direct animation of 3D avatars.




\subsection{Distillation-based Latent Coding}
\label{sec: decoder}

Our text-to-avatar generative model is based on the Latent Diffusion Model (LDM) framework~\cite{latentdiffusion}.  LDM generates images by reversing a noising process iteratively in a distilled latent space, reducing computational complexity while preserving semantic information. Recent works~\cite{direct3d, xiang2025structured, hu2024structldm, lan2024ln3diff} have validated its effectiveness on 3D generative tasks. In this paper, we pioneer the integration of UV-structured priors into text-to-avatar generation, thereby establishing the first LDM for text-to-3D avatar generation.  The typical training process of a text-conditioned LDM comprises two stages. A Variational Autoencoder (VAE) is trained in the first stage to establish a latent space. In the second stage, a text-conditioned diffusion model is trained to generate latent maps within this space, which are subsequently decoded by the trained decoder to produce the final results.  Through experimental observations, we found that directly training a VAE for complex 3D human models is prone to instability and demands substantial computational resources. Therefore, we propose a distillation-based decoding method that constructs the latent space based on a pre-trained, large-scale reconstruction model. This approach is not only robust but also requires significantly less computational resources.

Concretely, we leverage IDOL~\cite{zhuang2025idol}, a large reconstruction model with an encoder-decoder architecture, to encode the latent space.
IDOL directly reconstructs a 3D human model from a single input image and naturally constructs a generalizable and uniform feature space that maps the input image to the 3D human representation.
Its model consists of three main components: 1) The first part is a pre-trained high-resolution human foundation model, primarily responsible for capturing human poses and fine-grained appearance details from high-resolution human images. The output feature has a spatial resolution of 64×64 with 1536 channels. This feature space is aligned with the human image space and represents shallow features of the network, which are insufficient to capture 3D human structural information. 2) The second part is a UV-align transform, which aligns the previously obtained human image features into the UV feature space. The output consists of 9216 tokens, with a dimension of 1536. While this feature space contains rich 3D human structural and appearance information, its high dimensionality makes fitting its distribution using generative models challenging. 3) The third part is the UV Decoder, which converts the UV tokens from the previous part into a UV feature of size 1536×1536 with 32 channels. This UV feature is further decoded into 3D human Gaussians and rendered as an RGB image. 
Upon observation, this feature space exhibits good structural properties and relatively shallow feature representation close to final output, making it easier for the LDM to learn.

However, this UV feature cannot be directly utilized to train LDM due to its high resolution.
Therefore, we propose a distillation phase to construct a more compact representation from the original UV feature space.
Specifically, the uv feature maps from IDOL are down-sampled to a resolution of 256×256. 
A compact convolutional distillation decoder, consisting of upsampling and convolution operations, is then trained to restore these features to a 1024×1024 resolution.
Subsequently, two separate convolutional networks decode the geometry-related and color-related attributes of the 3D Gaussians into UV maps. Finally, the Gaussian attributes are obtained through UV sampling. The detailed network architecture of distillation network is provided \textit{in the supplementary material}.

These two networks form the distilled decoder, which reconstructs the 3D Gaussian human representation from the low-resolution latent feature. 
We randomly select four orthogonal views per avatar during training from the dataset as ground truth.
 Specifically, we input the front-view image into the IDOL encoder to obtain the corresponding UV features, which are then decoded by the distilled decoder into a 3D human. This 3D representation is rendered from the selected views, and the rendered results $I_{\text{pred}}$ are supervised by the corresponding ground truth images $I_{\text{gt}}$. 
The total training loss $ L_{\text{dist}}$ consists of two components: the image loss (L2 loss and a VGG loss) and the L2 regularization term for the Gaussian offsets:

\begin{align}
\nonumber
    L_{\text{dist}} = \sum_{i=1}^{N} \left( \| I_{\text{pred}} - I_{\text{gt}} \|^2 + \
\lambda_{\text{vgg}} L_{\text{vgg}}(I_{\text{pred}} - I_{\text{gt}}) \right)\\ 
+ \lambda_{\text{offset}} \| G_{\text{offset}} \|^2
\end{align}

\noindent where \(\lambda_{\text{L2}}\) = 20, \(\lambda_{\text{vgg}}\) = 20 and \(\lambda_{\text{offset}}\) = 1

\subsection{Structured Latent Diffusion Model}
\label{sec: diffusion}


We use Latent Diffusion to fit the structured UV latent distribution. To introduce text control, we employ CLIP as the text encoder and train the diffusion model using Classifier-Free Guidance. This enables the model to generate text-aligned structured UV features from noise.

\noindent\textbf{Text Conditioning.}
For the text annotations in the dataset, we follow Stable Diffusion and use CLIP\cite{radford2021learning} as the text encoder. After encoding the text with CLIP, we obtain 77 tokens of 768 channels, which are then mapped through a small linear layer and injected into the cross attention block of the diffusion model as the text condition.


\noindent\textbf{Classifier-free Guidance.}
We employ classifier-free guidance for text control. Expressly, during the training of the Diffusion model, we randomly set the text annotations to null for $20\%$ of the data, enabling joint training of both conditioned and unconditioned generation. During inference, the network outputs for the conditioned and unconditioned cases are linearly combined with a weight \( w \) to produce the final output~\cite{ho2021classifier}.




\noindent\textbf{Diffusion Model.}
Diffusion is a probabilistic model that fits the dataset distribution by progressively denoising Gaussian noise. The denoising process is an inverse discrete-time Markov chain of length \( T \). In the forward noising process, Gaussian noise \( e \sim \mathcal{N}(0, I) \) is gradually added to the samples from the dataset at each time step \( t \). At the \( t \)-th step, the noisy sample \( x_t \) is given by 

\begin{equation}
x_t := \alpha(t) x_0 + \sigma(t) e
\end{equation}

\noindent where both \( \alpha(t) \) and \( \sigma(t) \) are part of the noise scheduling. \( \alpha(t) \) represents the scaling factor applied to the original sample \( x_0 \) in the construction of the noisy sample \( x_t \), while \( \sigma(t) \) denotes the scaling factor of the noise at the current timestep. After \( T \) steps of noising, the sample is fully transformed into Gaussian noise. In the reverse denoising process, the Diffusion model starts with Gaussian noise at step \( T \) and progressively denoises until it recovers the clean sample \( x_0 \) at step 0.

We adopt an \( x_0 \)-prediction approach when training our Structured Latent Diffusion model. First, a feature \( f_0 \) is extracted from the structured latent feature as a sample. A time step \( t \) is randomly chosen from the range \( 1 \) to \( T \), and the corresponding \( \alpha_t \) and \( \sigma_t \) are generated using a noise scheduler. The feature \( f_t \) is then obtained using the equation for \( x_t \). The network is then tasked with predicting the clean feature \( \hat{f_0} \) corresponding to \( f_t \), and is supervised using the mean squared error (MSE) loss:

\begin{equation}
L_{\text{diff}} = \| \hat{f_0} - f_0 \|_2^2
\end{equation}


\subsection{Structure-Aware Editing}
\label{sec: edit}

Recently, SMPL-X-aligned approaches (e.g., IDOL~\cite{zhuang2025idol}) have facilitated texture editing of avatars by modifying the SMPL-X UV texture maps and controlling their shape via SMPL-X coefficients. However, due to the increased complexity of clothing geometry and texture, these methods generally struggle with tasks such as clothing replacement.

Benefiting from our effective distillation of IDOL, the latent space we obtain is exceptionally well-structured.
Consequently, editing the generated 3D avatars by manipulating the structured latent representation is straightforward. Nevertheless, because an avatar’s clothing is often strongly correlated with its identity, directly swapping the corresponding regions of the structured latent between two avatars can result in severe artifacts, such as unnatural edge transitions. Therefore, we choose to leverage the powerful inpainting capability of diffusion models to complete the regions of the structured latent corresponding to the clothing to be replaced, thereby producing a natural and plausible clothing swap effect.

Specifically, since we use latent diffusion to generate 3D digital humans and our latent space is well-structured, it is natural that we can perform virtual try-on by editing the structured latent through diffusion inpainting. Following Avrahami~\etal's work~\cite{avrahami2023blended}, we denote the latent corresponding to the 3D digital human as the background part $L_{bg}$, which needs to be preserved, and execute a specific denoising process to generate the modified foreground $L_{fg}$ as follows. First, we randomly sample the Gaussian noise to obtain the noise $L_{fg}^T$. At each denoising step t, we predict the noisy latent $L_{fg}^{t-1}$ for the previous step under the control of the target text, then add noise to the clean background latent $L_{bg}$ via the noise scheduler to get $L_{bg}^{t-1}$ for step $t-1$. By using a preprocessed foreground mask $mask_{fg}$, we combine $L_{bg}^{t-1}$ and $L_{fg}^{t-1}$ to obtain $L^{t-1}$. $L^{t-1}$ is then used as input for the network in the next denoising step to predict $L_{fg}^{t-2}$. This process is repeated until $t=0$, at which point the latent after the clothing change is obtained. The resulting latent is then passed into the decoder to generate the 3D human with the new clothing.



%% file: sec/4_exp.tex
\begin{figure*}[t]
    \centering
    \includegraphics[width=1.0\textwidth]{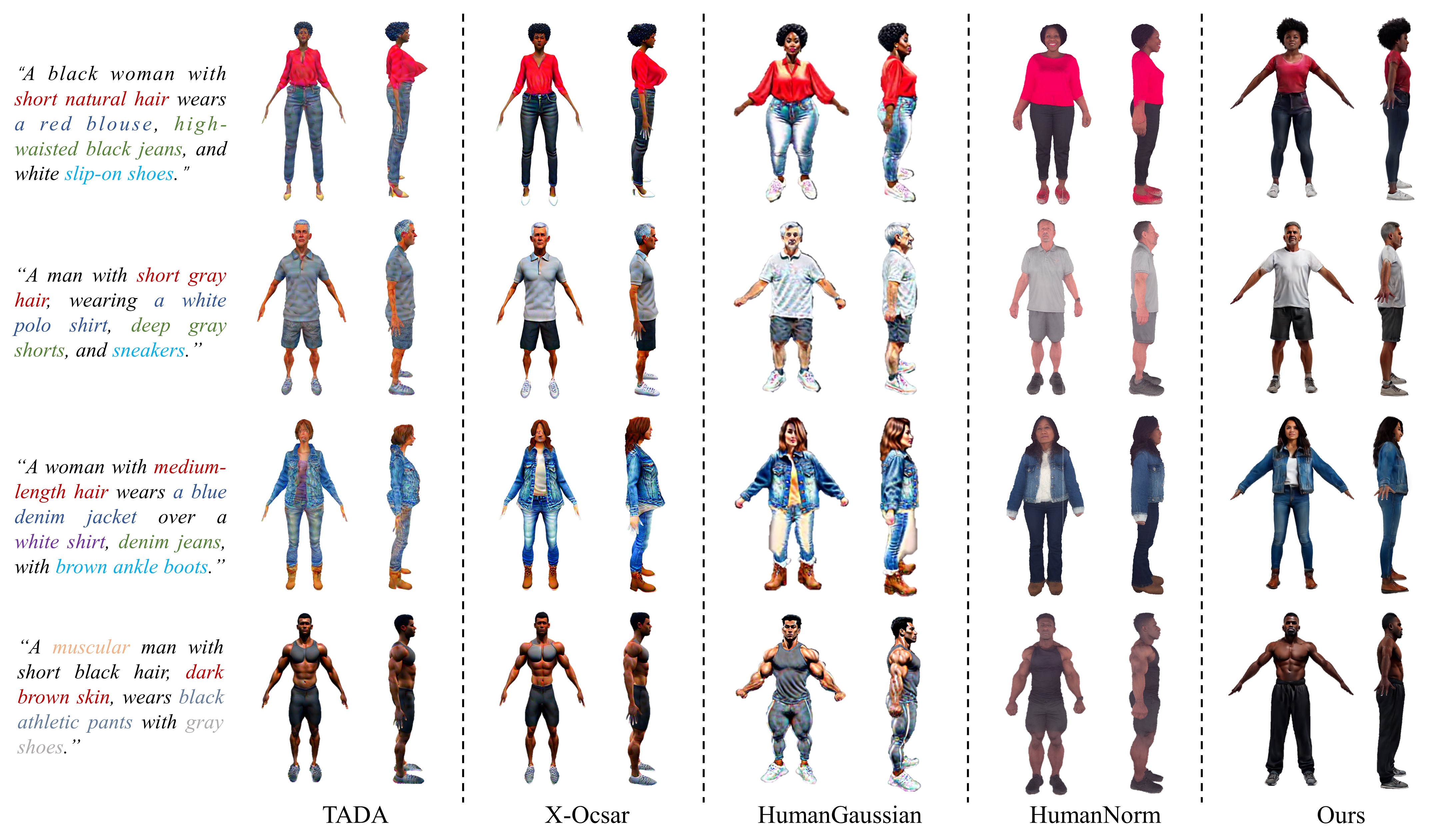}
    \vspace{-0.1in}
    \caption{Qualitative comparison of 3D avatars generated from four text prompts using our method and four baseline methods. The baselines often exhibit over-saturated colors, artifacts, and unrealistic body proportions. HumanNorm shows improved texture realism but struggles with accurate human proportions. Our method generates photorealistic avatars with natural textures and proper geometry.
    }
    \label{fig:comparison}
\end{figure*}

\section{Experiments}
\label{sec:exp}

\subsection{Implementation Details}
Our training set comprises 70,000 pairs of multi-view images, each annotated with SMPL-X parameters and accompanying text descriptions, as detailed in Sec.~\ref{sec: data}. For each individual, we utilize 24 views at a resolution of $896\times640$ for training, with text descriptions ranging from 12 to 40 words.

The distillation-based latent decoder is trained on 4 NVIDIA RTX A6000 GPU with a batch size of 2. For each person, four orthogonal views are randomly selected for both rendering and supervisory signals.  The structured latent diffusion model is trained on 4 NVIDIA RTX 3090 GPU with a batch size of 8, utilizing the DDPM noise scheduler with 1000 steps.  The entire training process takes approximately 90 hours to complete.
In the inference phase, we apply the DDPM sampling method with 100 denoising steps to refine latent noise into a 3D representation. The output is then decoded by the distillation-based latent decode, completing the process in ~12 seconds on an NVIDIA RTX 3090 GPU.

\subsection{Text-Guided 3D Human Generation}
\noindent\textbf{Baselines.}
We compare our proposed method with existing state-of-the-art text-to-3D human generation methods, including TADA~\cite{liao2024tada}, X-oscar~\cite{ma2024x}, HumanGaussian~\cite{liu2024humangaussian}, and HumanNorm~\cite{huang2023humannorm}. All these baseline methods are SDS-based 3D avatar generative models.

\noindent\textbf{Qualitative comparison.} The rendered results of all methods are shown in Fig.~\ref{fig:comparison}. Under prompts with everyday clothing, SDS-based methods generally fail to generate realistic avatars. Specifically, HumanGaussian, TADA, and X-OSCAR, which directly distill Stable Diffusion using SDS Loss, exhibit overly saturated and unrealistic colors. Furthermore, due to the lack of real human geometric supervision, TADA and X-OSCAR produce avatars with disproportionately small heads, thin arms, and overly long legs. HumanGaussian generates flat and disproportionate human figures.  
HumanNorm introduces Normal Diffusion and Multi-step SDS Loss, partially mitigating body proportions and color oversaturation issues. However, due to the inherent bias of SDS Loss, discrepancies remain between the distilled knowledge and realistic human distributions, leading to artifacts in the face, forearms, hands, and feet. 

In contrast, our method directly learns the distribution of real human bodies using diffusion, avoiding the issues of color oversaturation and unrealistic geometry in other methods. 
Additionally, our single-pass generation process leverages the diffusion denoising process without iterative optimization, achieving significantly higher efficiency with an inference time of 12s on an NVIDIA RTX 3090 GPU, compared to several hours required by the baseline methods.

\noindent\textbf{Quantitative comparison.} 
We adopt the CLIP Score \cite{hessel2021clipscore}, VQA Score and user study to evaluate the five methods objectively and subjectively. The test prompts for evaluation are generated by ChatGPT with random appearance, and the results are reported in Tab.~\ref{table:quantitative_comparison}

For objective comparison, CLIP score is leveraged to assess the consistency between the input text description and the output renderings.  Our method achieves the second-highest CLIP Score among all approaches. We further adopted the VQA score with Qwen  from Progressive3D \cite{chengprogressive3d}. Each text prompt is decomposed into a small set of yes/no questions, and a vision-language model predicts the proportion of “yes” answers as a consistency score. When evaluated on our benchmark prompts in Tab.~\ref{table:quantitative_comparison}, TeRA achieves the best performance. This demonstrates that TeRA strong adherence to textual descriptions. This observation is further corroborated by user study results, which show that our method received the highest preference score, demonstrating exceptional realism and alignment with textual descriptions.

For subjective comparison, we invited 28 participants to evaluate different methods based on three criteria: text consistency (Tex.), visual quality (Vis.), and realism (Real.), using questionnaire rating from 0 to 5.  The results reveal that the TeRA model excels by achieving the highest score across all three questions, markedly surpassing the runner-up, thereby demonstrating our model's superior text-appearance consistency, enhanced realism, and improved rendering quality.

The runtimes of different methods are also reported in Table~\ref {table:quantitative_comparison}.  As all four other methods are SDS-based methods requiring iterative optimization for each generation, their runtimes are typically more than 1 hour on a single Nvidia RTX 3090 GPU. In contrast, our method boasts a single generation time of just 12 seconds, representing a significant improvement of two orders of magnitude in speed compared to other methods.

\begin{table}[]
\centering
\resizebox{\columnwidth}{!}{
\begin{tabular}{@{}lcccccc@{}}
\toprule
\multirow{2}{*}{Method} & CLIP & VQA & \multicolumn{3}{c}{User Study}  & \multirow{2}{*}{Time↓} \\ \cmidrule(lr){4-6}
                                 & Score↑    &  Score↑ & \multicolumn{1}{l}{Tex.↑} & \multicolumn{1}{l}{Vis.↑} & \multicolumn{1}{l}{Real.↑} &  \\ \midrule
TADA~\cite{liao2024tada}   & 29.86       & 0.64      & 3.27         & 2.25              & 2.11              & 2.3h              \\
X-Oscar~\cite{ma2024x}     & {\ul \textbf{32.46}} & 0.80 & 3.56         & 2.54              & 2.26              & 2.0h               \\
HumanGaussiann~\cite{liu2024humangaussian}       & 29.31     & {\ul \textbf{0.82}}       & 3.74              & 2.49              & 2.28          & {\ul 1.0h}\\
HumanNorm~\cite{huang2023humannorm}               & 29.94     & 0.72      & {\ul 3.79}        & {\ul 3.01}        & {\ul 3.04}    & 4.0h           \\
TeRA (Ours)     & {\ul 30.17}        & {\ul \textbf{0.82}}   & {\ul \textbf{4.54}}  & {\ul \textbf{4.33}}      & {\ul \textbf{4.35}}      & {\ul \textbf{12s}}\\ \bottomrule
\end{tabular}
}
\vspace{-0.1in}
 \caption{Quantitative comparison of CLIP Score, VQA Score, User Study results and time cost for text-to-3D human generation methods. The {\ul \textbf{best}} and {\ul second-best} scores are marked.}
\label{table:quantitative_comparison}
\end{table}

\subsection{Ablation Study}
\begin{figure}[t]
    \centering
    \includegraphics[width=1\linewidth]{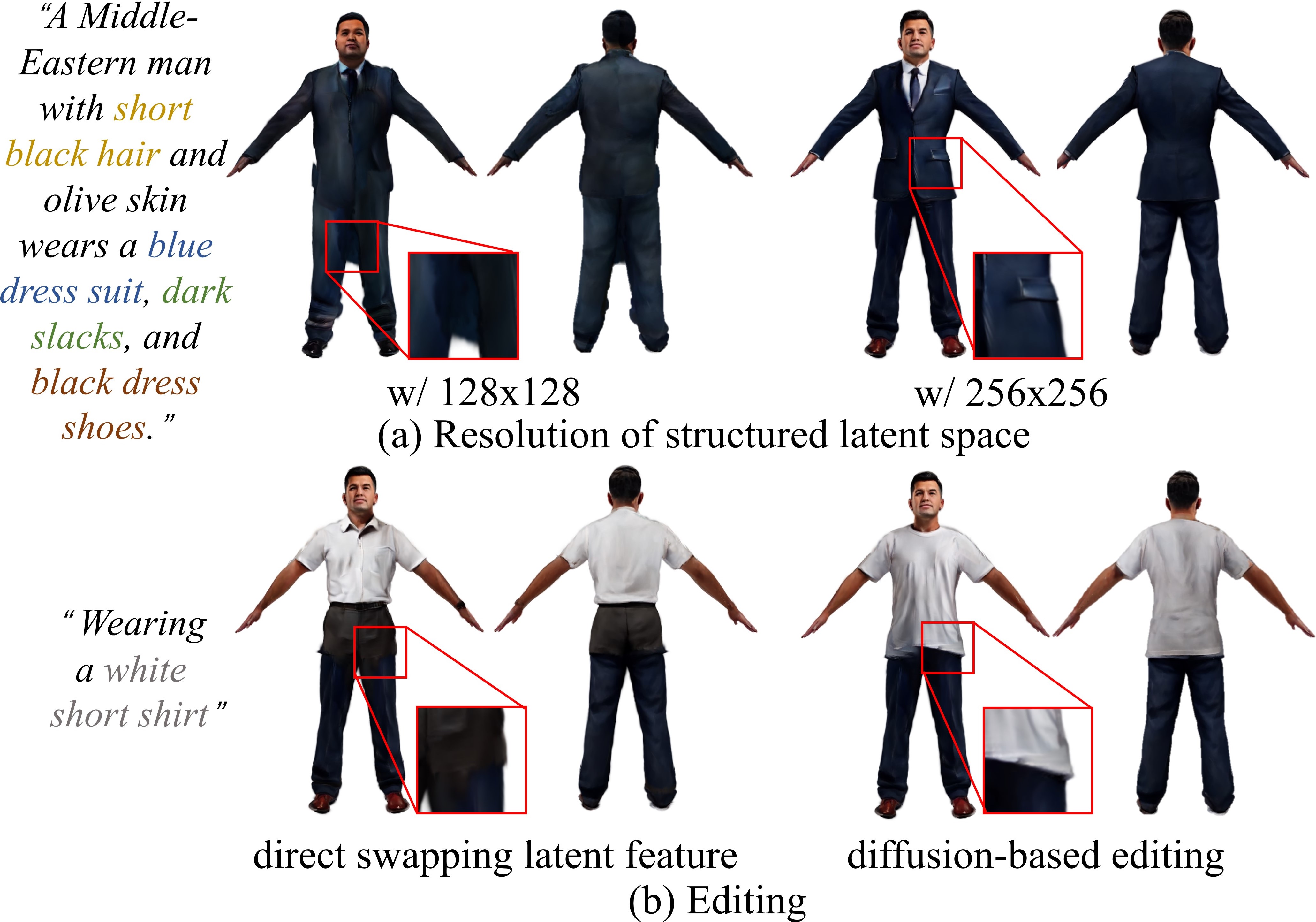}
    \caption{Ablation study on key components of TeRA. (a) Comparison of structured latent space resolutions ($128 \times 128$ vs. $256 \times 256$) showing that higher resolution provides richer details and fewer artifacts. (b) Evaluation of the proposed inpainting method for virtual try-on, demonstrating that inpainting on the structured latent space yields smoother transitions and fewer artifacts compared to direct feature swapping. 
    }
    \vspace{-0.1in}
    \label{fig:ablation}
\end{figure}

As illustrated in Fig.~\ref{fig:ablation}, we perform an ablation study focusing on two key modules: the resolution of the distilled structured latent space and our novel inpainting strategy designed for virtual try-on.


\textbf{Latent Space Resolution.} In Fig.~\ref{fig:ablation}-(a), we compare the performance of structured latent space with resolutions of $128 \times 128$ and $256 \times 256$. The results demonstrate that the higher resolution of $256 \times 256$ provides richer details and significantly reduces artifacts in the generated 3D avatars. The results validate our choice of a $256 \times 256$ resolution as an optimal balance, providing high-quality outputs without excessively increasing the training cost.

\textbf{Inpainting Strategy.} We evaluate the effectiveness of our proposed inpainting method for virtual try-on, as shown in Fig.~\ref{fig:ablation}-(b). We compare our method to a baseline approach that directly swaps the latent features in the specified region with newly generated features guided by the new prompt. As evidenced in the figure, our proposed inpainting technique applied to the structured latent space results in smoother transitions and significantly reduces artifacts compared to the direct feature swapping method.

\subsection{Application}
\begin{figure}[t]
    \centering
    \includegraphics[width=0.95\linewidth]{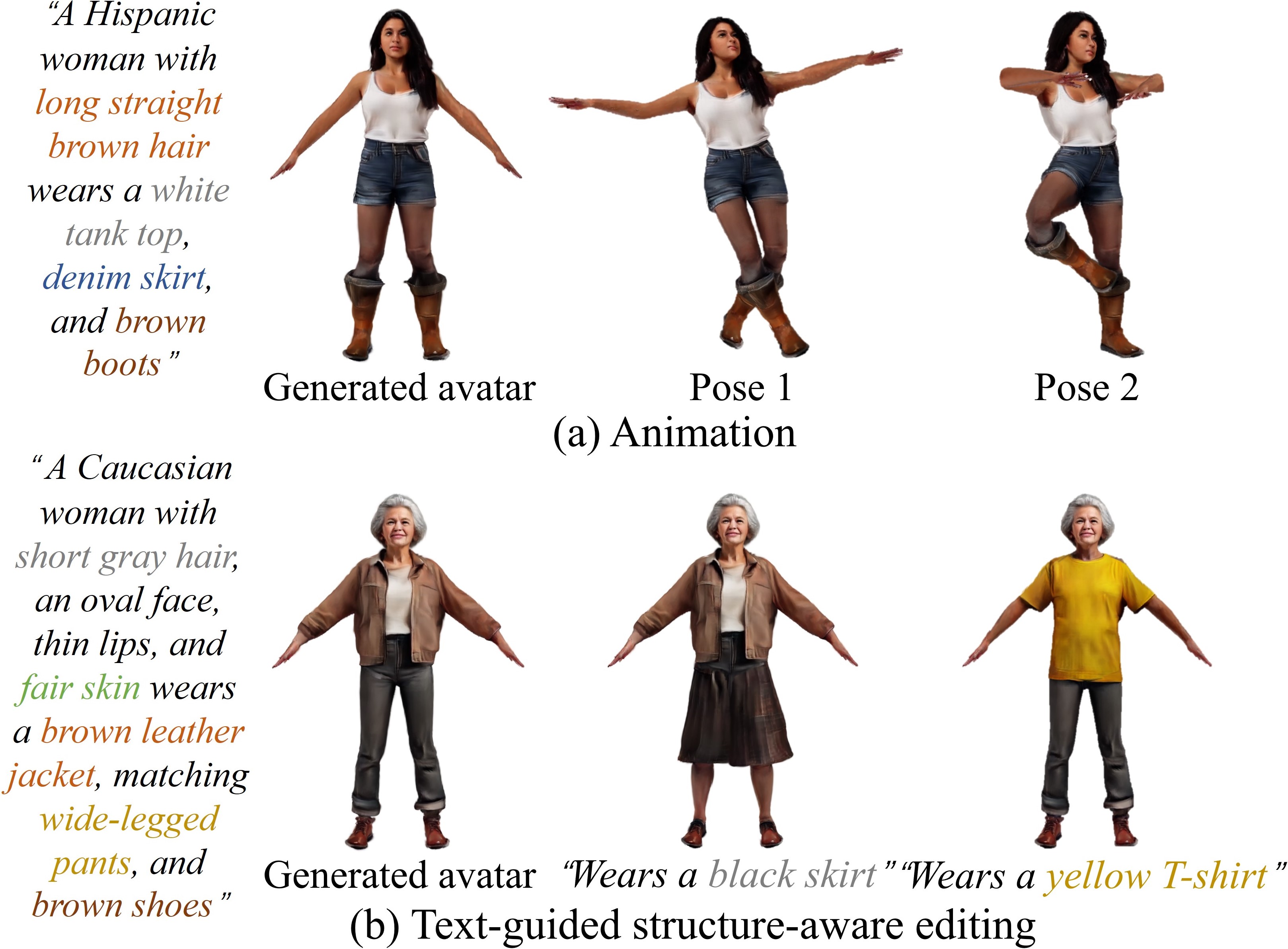}
    \caption{Illustration of TeRA's downstream applications. The upper row shows direct avatar animation using SMPL-X poses without post-processing, while the lower row presents natural virtual try-on results via diffusion-based latent space editing.
    }
    \label{fig:application}
\end{figure}

Our proposed model generates 3D avatars by learning structured latent representations through diffusion, enabling versatile downstream applications such as \textit{editing} and \textit{animation}, as illustrated in Fig.~\ref{fig:application}. 
Since our method directly generates the structured latent representation using diffusion, it supports inpainting operations on the generated latent space, allowing seamless 3D avatar editing such as virtual try-on. 
Additionally, representing the 3D human using a combination of SMPL-X and Gaussian Attribute Maps enables flexible texture editing through color map modifications and shape editing by altering SMPL-X parameters. 
Furthermore, this design facilitates straightforward animation by directly driving the generated avatars using SMPL-X pose sequences, eliminating the need for post-processing and ensuring efficient and realistic motion control.

%% file: sec/5_con.tex
\section{Conclusion}
\label{sec:con}
We introduce TeRA, a text-to-3D avatar generation model that achieves fast and high-quality 3D human reconstruction. By leveraging a structured latent space through distilling a pre-trained large reconstruction model, TeRA produces photorealistic avatars with strong text-model alignment, outperforming state-of-the-art SDS-based models in both speed and visual quality. 

TeRA still faces certain limitations. As the training data consists of static models, it cannot model dynamic details like clothing wrinkles resulting from human movement. Furthermore, due to TeRA's reliance on the SMPL-X model for human body representation, its modeling quality is limited for loose garments such as dresses.

%% file: sec/X_suppl.tex
\maketitlesupplementary

\begin{figure}[b]
    \centering
    \includegraphics[width=0.95\linewidth]{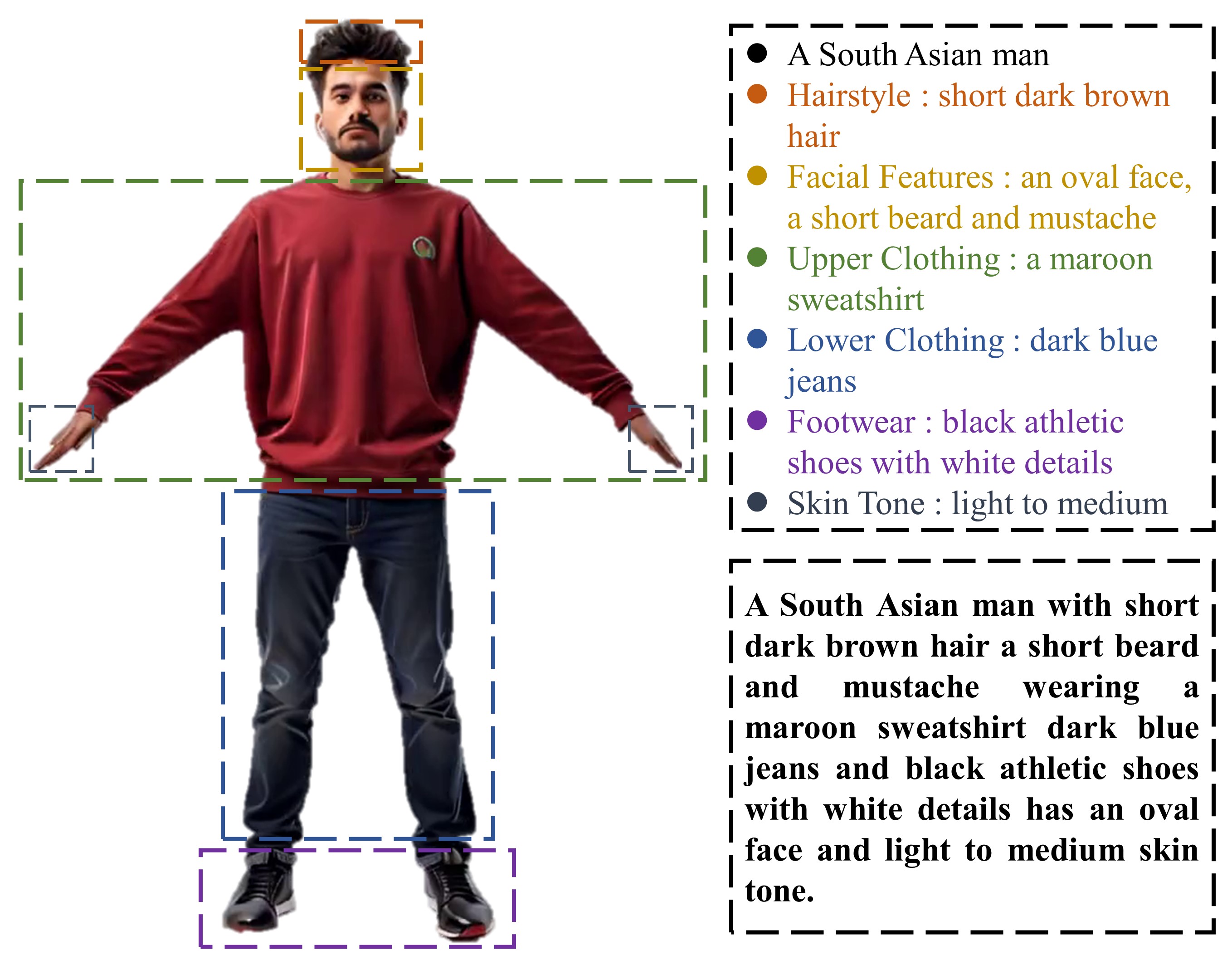}
    \caption{Details of our dataset
    }
    \label{fig:dataset}
\end{figure}

\begin{figure}[]
    \centering
    \includegraphics[width=0.95\linewidth]{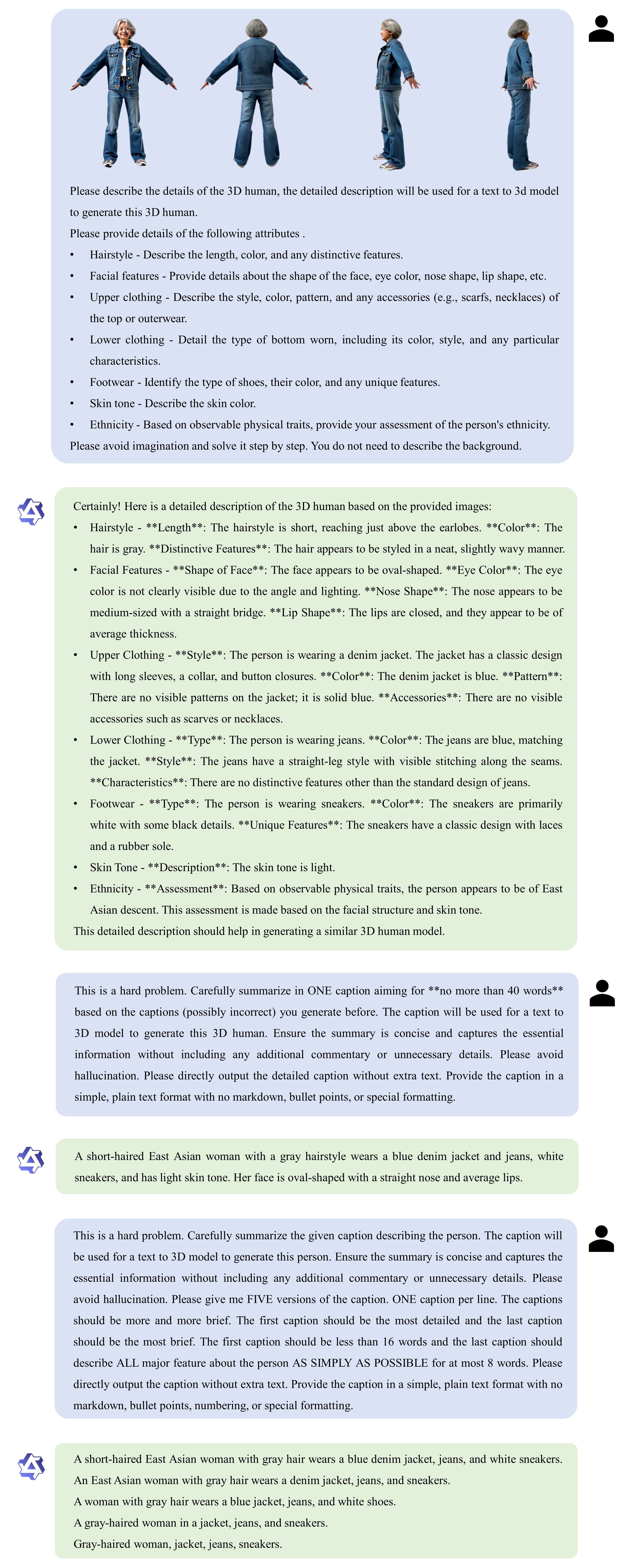}
    \caption{An example of our captioning process.
    }
    \label{fig:data_caption}
\end{figure}

\section{Dataset Details and Comparisons}
A sample of our dataset is detailed in Fig.~\ref{fig:dataset}.And as shown in Tab.~\ref{table:comparsion_with_other_dataset}, our dataset is the largest text-annotated multi-view human dataset to date, containing significantly more identities than MVHumanNet—the only comparable dataset with textual annotations. Moreover, our annotations are more accurate and comprehensive than previous datasets.

\begin{table}[h]
\scriptsize
\centering
\resizebox{1\linewidth}{!}{
\begin{tabular}{@{}lcccc@{}}
\toprule
\textbf{Dataset}       & \textbf{Frames} & \textbf{ID}    & \textbf{View} & \textbf{Text Caption} \\ \midrule
HuMMan        & 60M      & 1000                     & 10     & \ding{55}              \\
HUMBI         & 26M      & 772                      & 107    & \ding{55}              \\
DNA-Rendering & 67.5M    & 500                      & 60     & \ding{55}              \\
MVHumanNet    & 645.1M   & 4500                     & 48     & \ding{51}              \\ \midrule
THuman2.1     & -        & 2500                     & Free   & \ding{55}              \\
2K2K          & -        & \multicolumn{1}{l}{2050} & Free   & \ding{55}              \\ \midrule
Ours          & 2.4M     & 100K                     & 24     & \ding{51}              \\ \bottomrule
\end{tabular}
}
\caption{Comparisons of datasets.}
\label{table:comparsion_with_other_dataset}
\end{table}

\section{Data Caption}
\label{sec:rationale}
Since it is challenging for a large language model to output accurate labels of varying lengths in a single conversation, we annotate text over three rounds of dialogue. Precisely, to capture as much information as possible from the input images, we prompt the Qwen-2.5VL model in the first round to provide detailed descriptions of four orthogonal views of the human body, including various pertinent details. In this round, Qwen-2.5VL outputs a comprehensive description; however, to meet the input text length requirements of the CLIP model, these verbose descriptions must be condensed. In the second round, we feed the output from the first round into the Qwen2.5 model and request that it summarize the detailed description into a long annotation of no more than 40 words. Finally, to further enrich the textual content, we ask Qwen2.5 to further condense the outputs from the first two rounds into five descriptions of varying lengths, with the longest being no more than 16 words and the shortest containing at least 8 words. During training, one of these five short descriptions or the 40-word long annotation is randomly selected as the condition.The complete text annotation process is illustrated in Fig.~\ref{fig:data_caption}.

\section{Architecture of Distillation Decoder}
The distillation decoder consists of a UV code decoder and a Gaussian attribute decoding head. The UV code decoder includes two transposed convolution layers and two convolution layers, with an input feature size of 256 × 256 × 32 and an output code size of 1024 × 1024 × 32. The output code from the UV encoder-decoder is split into two parts: the first 16 channels are regarded as the geometry code, and the remaining 16 channels as the texture code. The Gaussian attribute decoding head consists of three convolutional heads, each comprising two or three convolutional layers. These Gaussian attribute decoding heads are responsible for decoding the geometry code and texture code into five 3D Gaussian attributes in the SMPL-X texture space.

\begin{figure}[!h]
    \centering
    \includegraphics[width=1\linewidth]{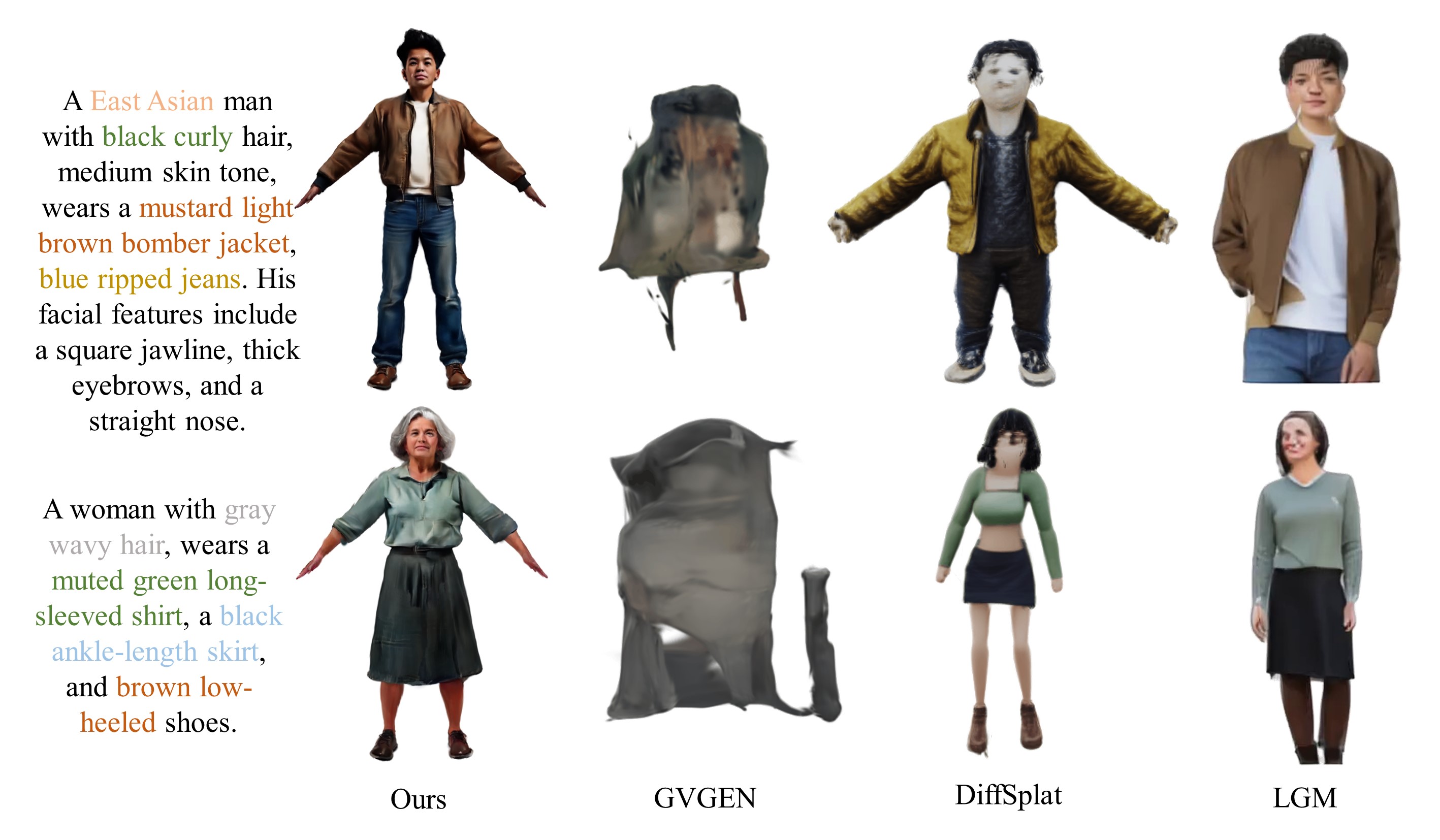}
    \caption{Additional qualitative comparisons with general 3D methods.}
    \label{fig:comp_with_general3d}
\end{figure}

\section{More Comparisons}
As far as we know, SDS-based models are the only available text-to-3D-avatar methods.  We have evaluated additional baselines of general 3D reconstruction methods, including LGM (text → multi-view → 3DGS), GVGen (direct 3D), and DiffSplat (2D‐diffusion → 3D) in Fig.~\ref{fig:comp_with_general3d}; all deliver lower visual fidelity and weaker prompt adherence than TeRA. Our representation simplifies the learning of the target distribution and enables downstream applications.

\section{More Results}

\begin{figure*}[t]
    \centering
    \includegraphics[width=0.95\linewidth]{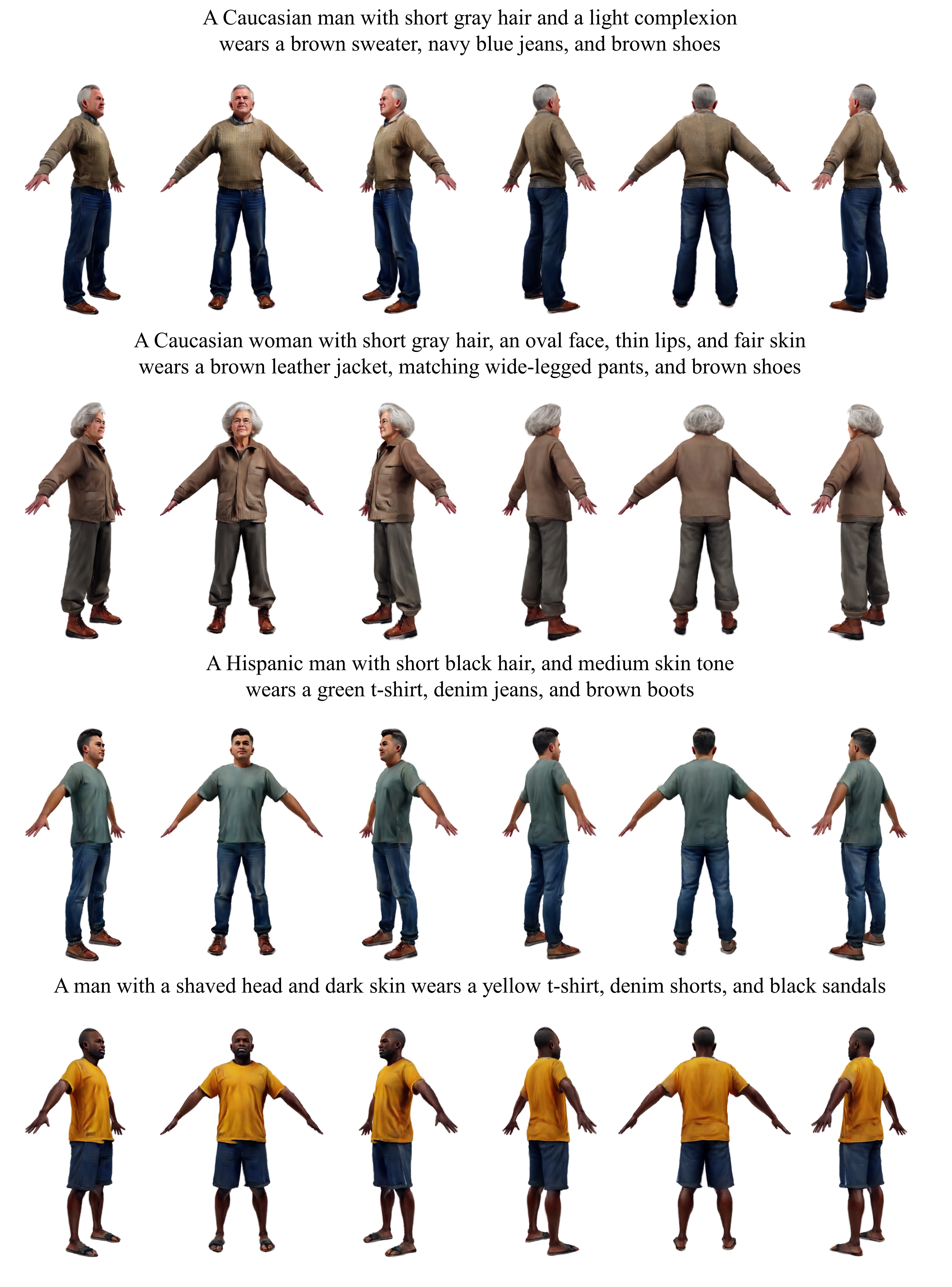}
    \caption{More results of text-guided generation
    }
    \label{fig:more_results1}
\end{figure*}

\begin{figure*}[t]
    \centering
    \includegraphics[width=0.95\linewidth]{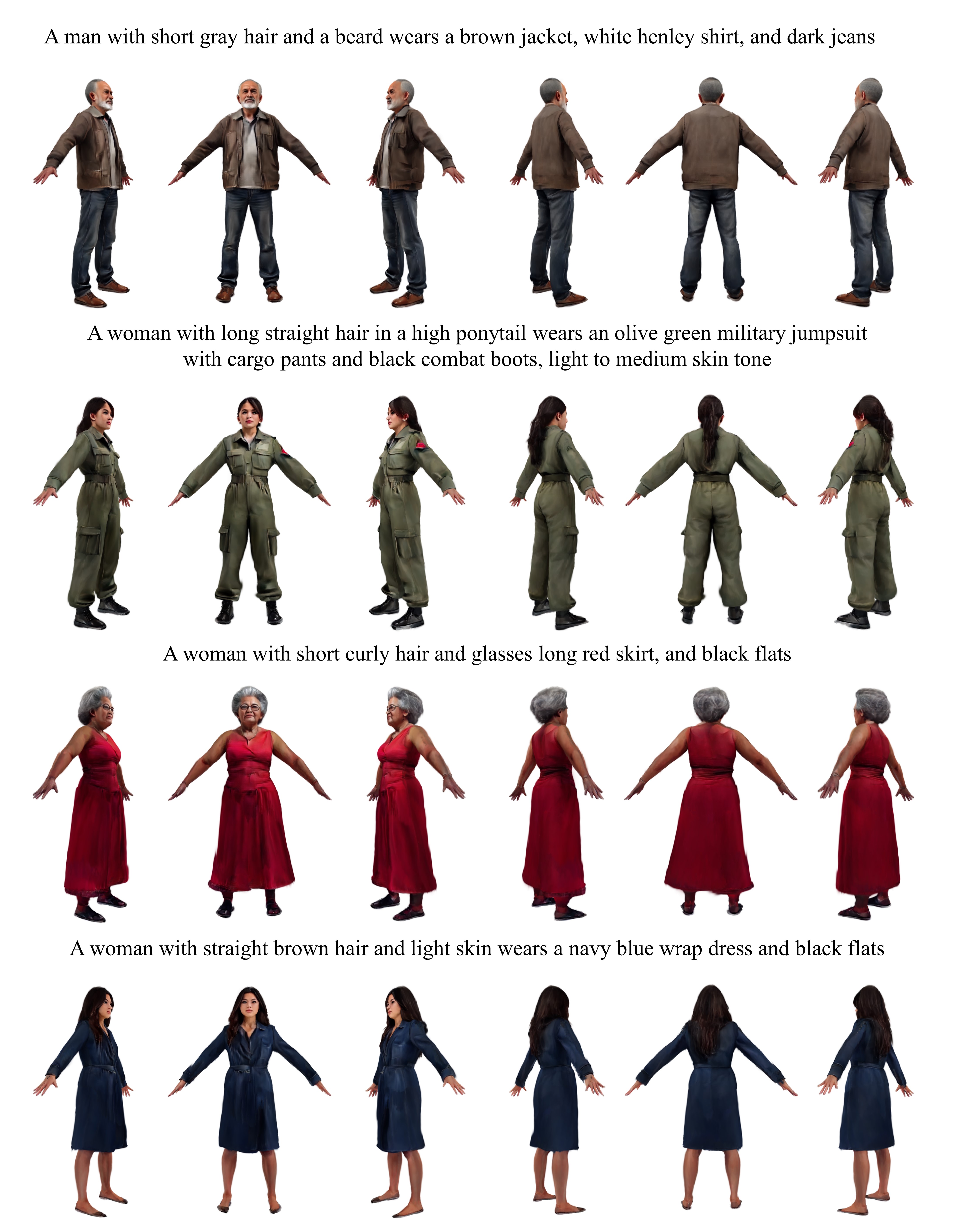}
    \caption{More results of text-guided generation
    }
    \label{fig:more_results2}
\end{figure*}

We show more renderings of our generated models with input text description in Fig.~\ref{fig:more_results1} and Fig.~\ref{fig:more_results2}.

\begin{figure*}[t]
    \centering
    \includegraphics[width=0.95\linewidth]{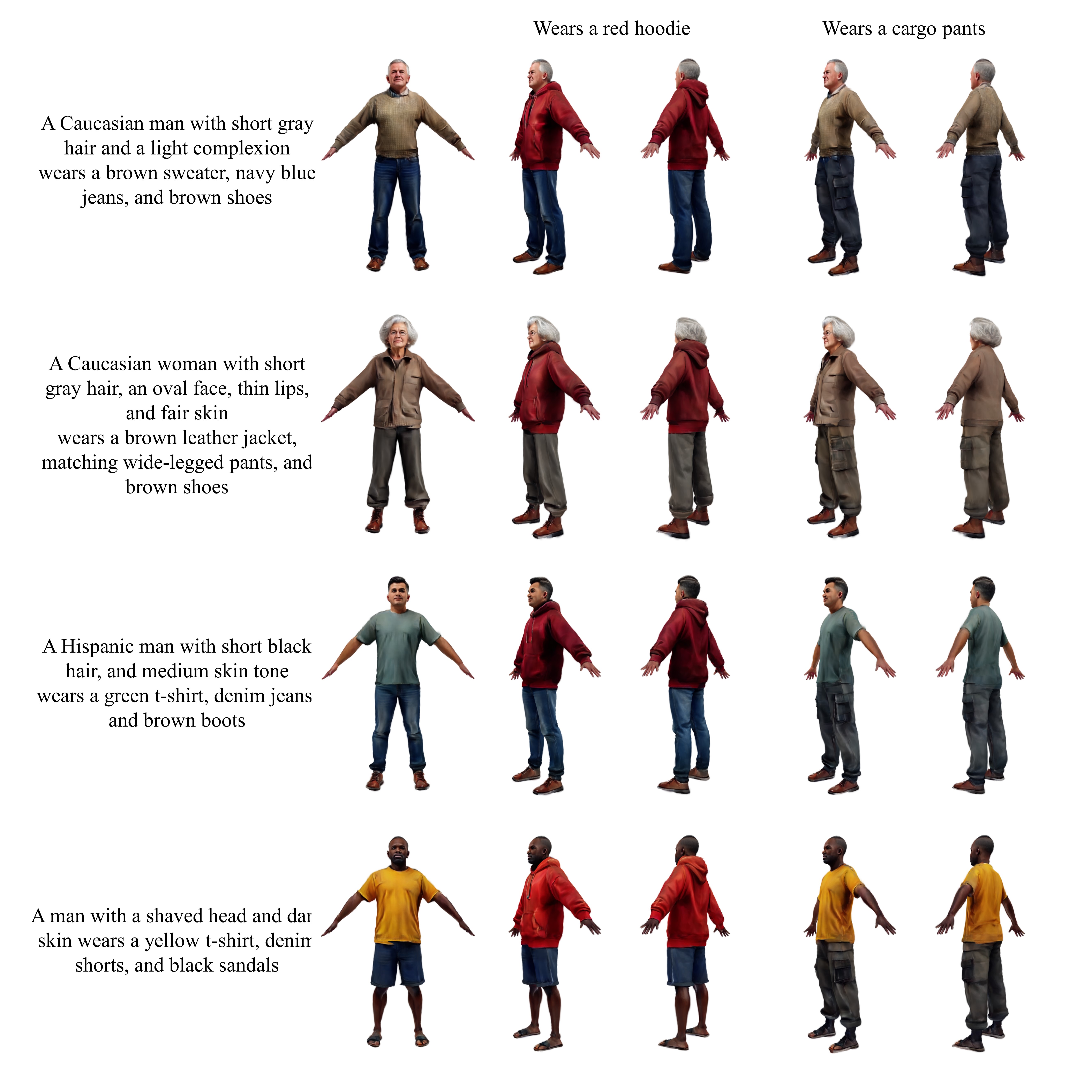}
    \caption{More results of text-guided virtual try-on
    }
    \label{fig:more_results_3}
\end{figure*}

%% file: main.bbl
\begin{thebibliography}{98}
\providecommand{\natexlab}[1]{#1}
\providecommand{\url}[1]{\texttt{#1}}
\expandafter\ifx\csname urlstyle\endcsname\relax
  \providecommand{\doi}[1]{doi: #1}\else
  \providecommand{\doi}{doi: \begingroup \urlstyle{rm}\Url}\fi

\bibitem[Achiam et~al.(2023)Achiam, Adler, Agarwal, Ahmad, Akkaya, Aleman, Almeida, Altenschmidt, Altman, Anadkat, et~al.]{achiam2023gpt}
Josh Achiam, Steven Adler, Sandhini Agarwal, Lama Ahmad, Ilge Akkaya, Florencia~Leoni Aleman, Diogo Almeida, Janko Altenschmidt, Sam Altman, Shyamal Anadkat, et~al.
\newblock Gpt-4 technical report.
\newblock \emph{arXiv preprint arXiv:2303.08774}, 2023.

\bibitem[Achlioptas et~al.(2018)Achlioptas, Diamanti, Mitliagkas, and Guibas]{achlioptas2018learning}
Panos Achlioptas, Olga Diamanti, Ioannis Mitliagkas, and Leonidas Guibas.
\newblock Learning representations and generative models for 3d point clouds.
\newblock In \emph{International conference on machine learning}, pages 40--49. PMLR, 2018.

\bibitem[Alexander et~al.(2010)Alexander, Rogers, Lambeth, Chiang, Ma, Wang, and Debevec]{alexander2010digital}
Oleg Alexander, Mike Rogers, William Lambeth, Jen-Yuan Chiang, Wan-Chun Ma, Chuan-Chang Wang, and Paul Debevec.
\newblock The digital emily project: Achieving a photorealistic digital actor.
\newblock \emph{IEEE Computer Graphics and Applications}, 30\penalty0 (4):\penalty0 20--31, 2010.

\bibitem[Alldieck et~al.(2021)Alldieck, Xu, and Sminchisescu]{imGHUM}
Thiemo Alldieck, Hongyi Xu, and Cristian Sminchisescu.
\newblock imghum: Implicit generative models of 3d human shape and articulated pose.
\newblock In \emph{ICCV}, 2021.

\bibitem[An et~al.(2023)An, Xu, Shi, Song, Ogras, and Luo]{an2023panohead}
Sizhe An, Hongyi Xu, Yichun Shi, Guoxian Song, Umit~Y Ogras, and Linjie Luo.
\newblock Panohead: Geometry-aware 3d full-head synthesis in 360deg.
\newblock In \emph{Proceedings of the IEEE/CVF conference on computer vision and pattern recognition}, pages 20950--20959, 2023.

\bibitem[Asmussen and Taksar(1997)]{asmussen1997controlled}
S{\o}ren Asmussen and Michael Taksar.
\newblock Controlled diffusion models for optimal dividend pay-out.
\newblock \emph{Insurance: Mathematics and Economics}, 20\penalty0 (1):\penalty0 1--15, 1997.

\bibitem[Avrahami et~al.(2023)Avrahami, Fried, and Lischinski]{avrahami2023blended}
Omri Avrahami, Ohad Fried, and Dani Lischinski.
\newblock Blended latent diffusion.
\newblock \emph{ACM transactions on graphics (TOG)}, 42\penalty0 (4):\penalty0 1--11, 2023.

\bibitem[Bai et~al.(2025)Bai, Chen, Liu, Wang, Ge, Song, Dang, Wang, Wang, Tang, et~al.]{bai2025qwen2}
Shuai Bai, Keqin Chen, Xuejing Liu, Jialin Wang, Wenbin Ge, Sibo Song, Kai Dang, Peng Wang, Shijie Wang, Jun Tang, et~al.
\newblock Qwen2. 5-vl technical report.
\newblock \emph{arXiv preprint arXiv:2502.13923}, 2025.

\bibitem[Balaji et~al.(2022)Balaji, Nah, Huang, Vahdat, Song, Kreis, Aittala, Aila, Laine, Catanzaro, et~al.]{balaji2022ediffi}
Yogesh Balaji, Seungjun Nah, Xun Huang, Arash Vahdat, Jiaming Song, Karsten Kreis, Miika Aittala, Timo Aila, Samuli Laine, Bryan Catanzaro, et~al.
\newblock ediffi: Text-to-image diffusion models with an ensemble of expert denoisers.
\newblock \emph{arXiv preprint arXiv:2211.01324}, 2022.

\bibitem[Bansal et~al.(2023)Bansal, Chu, Schwarzschild, Sengupta, Goldblum, Geiping, and Goldstein]{bansal2023universal}
Arpit Bansal, Hong-Min Chu, Avi Schwarzschild, Soumyadip Sengupta, Micah Goldblum, Jonas Geiping, and Tom Goldstein.
\newblock Universal guidance for diffusion models.
\newblock In \emph{Proceedings of the IEEE/CVF Conference on Computer Vision and Pattern Recognition}, pages 843--852, 2023.

\bibitem[Blattmann et~al.(2023)Blattmann, Dockhorn, Kulal, Mendelevitch, Kilian, Lorenz, Levi, English, Voleti, Letts, et~al.]{blattmann2023stable}
Andreas Blattmann, Tim Dockhorn, Sumith Kulal, Daniel Mendelevitch, Maciej Kilian, Dominik Lorenz, Yam Levi, Zion English, Vikram Voleti, Adam Letts, et~al.
\newblock Stable video diffusion: Scaling latent video diffusion models to large datasets.
\newblock \emph{arXiv preprint arXiv:2311.15127}, 2023.

\bibitem[Cao et~al.(2024)Cao, Cao, Han, Shan, and Wong]{cao2024dreamavatar}
Yukang Cao, Yan-Pei Cao, Kai Han, Ying Shan, and Kwan-Yee~K Wong.
\newblock Dreamavatar: Text-and-shape guided 3d human avatar generation via diffusion models.
\newblock In \emph{Proceedings of the IEEE/CVF conference on computer vision and pattern recognition}, pages 958--968, 2024.

\bibitem[Chan et~al.(2022)Chan, Lin, Chan, Nagano, Pan, De~Mello, Gallo, Guibas, Tremblay, Khamis, et~al.]{chan2022efficient}
Eric~R Chan, Connor~Z Lin, Matthew~A Chan, Koki Nagano, Boxiao Pan, Shalini De~Mello, Orazio Gallo, Leonidas~J Guibas, Jonathan Tremblay, Sameh Khamis, et~al.
\newblock Efficient geometry-aware 3d generative adversarial networks.
\newblock In \emph{Proceedings of the IEEE/CVF Conference on Computer Vision and Pattern Recognition}, pages 16123--16133, 2022.

\bibitem[Chen et~al.(2019)Chen, Choy, Savva, Chang, Funkhouser, and Savarese]{chen2019text2shape}
Kevin Chen, Christopher~B Choy, Manolis Savva, Angel~X Chang, Thomas Funkhouser, and Silvio Savarese.
\newblock Text2shape: Generating shapes from natural language by learning joint embeddings.
\newblock In \emph{Computer Vision--ACCV 2018: 14th Asian Conference on Computer Vision, Perth, Australia, December 2--6, 2018, Revised Selected Papers, Part III 14}, pages 100--116. Springer, 2019.

\bibitem[Chen et~al.(2023)Chen, Chen, Jiao, and Jia]{chen2023fantasia3d}
Rui Chen, Yongwei Chen, Ningxin Jiao, and Kui Jia.
\newblock Fantasia3d: Disentangling geometry and appearance for high-quality text-to-3d content creation.
\newblock In \emph{Proceedings of the IEEE/CVF international conference on computer vision}, pages 22246--22256, 2023.

\bibitem[Chen et~al.(2024)Chen, Wang, Wang, and Liu]{chen2024text}
Zilong Chen, Feng Wang, Yikai Wang, and Huaping Liu.
\newblock Text-to-3d using gaussian splatting.
\newblock In \emph{Proceedings of the IEEE/CVF conference on computer vision and pattern recognition}, pages 21401--21412, 2024.

\bibitem[Cheng et~al.(2019)Cheng, Bronstein, Zhou, Kotsia, Pantic, and Zafeiriou]{cheng2019meshgan}
Shiyang Cheng, Michael Bronstein, Yuxiang Zhou, Irene Kotsia, Maja Pantic, and Stefanos Zafeiriou.
\newblock Meshgan: Non-linear 3d morphable models of faces.
\newblock \emph{arXiv preprint arXiv:1903.10384}, 2019.

\bibitem[Cheng et~al.()Cheng, Yang, Wang, Li, Zhang, Zhang, and Yuan]{chengprogressive3d}
Xinhua Cheng, Tianyu Yang, Jianan Wang, Yu Li, Lei Zhang, Jian Zhang, and Li Yuan.
\newblock Progressive3d: Progressively local editing for text-to-3d content creation with complex semantic prompts.
\newblock In \emph{The Twelfth International Conference on Learning Representations}.

\bibitem[Cheong et~al.(2024)Cheong, Ceylan, Mustafa, Gilbert, and Huang]{cheong2024boosting}
Soon~Yau Cheong, Duygu Ceylan, Armin Mustafa, Andrew Gilbert, and Chun-Hao~Paul Huang.
\newblock Boosting camera motion control for video diffusion transformers.
\newblock \emph{arXiv preprint arXiv:2410.10802}, 2024.

\bibitem[Debevec(2012)]{debevec2012light}
Paul Debevec.
\newblock The light stages and their applications to photoreal digital actors.
\newblock \emph{SIGGRAPH Asia}, 2\penalty0 (4):\penalty0 1--6, 2012.

\bibitem[Gao et~al.(2023)Gao, Aigerman, Groueix, Kim, and Hanocka]{gao2023textdeformer}
William Gao, Noam Aigerman, Thibault Groueix, Vova Kim, and Rana Hanocka.
\newblock Textdeformer: Geometry manipulation using text guidance.
\newblock In \emph{ACM SIGGRAPH 2023 Conference Proceedings}, pages 1--11, 2023.

\bibitem[Ge et~al.(2024)Ge, Zeng, Huffman, Lin, Liu, and Cui]{ge2024visual}
Yunhao Ge, Xiaohui Zeng, Jacob~Samuel Huffman, Tsung-Yi Lin, Ming-Yu Liu, and Yin Cui.
\newblock Visual fact checker: Enabling high-fidelity detailed caption generation.
\newblock In \emph{Proceedings of the IEEE/CVF Conference on Computer Vision and Pattern Recognition}, pages 14033--14042, 2024.

\bibitem[G{\"u}ler et~al.(2018)G{\"u}ler, Neverova, and Kokkinos]{guler2018densepose}
R{\i}za~Alp G{\"u}ler, Natalia Neverova, and Iasonas Kokkinos.
\newblock Densepose: Dense human pose estimation in the wild.
\newblock In \emph{Proceedings of the IEEE conference on computer vision and pattern recognition}, pages 7297--7306, 2018.

\bibitem[Guo et~al.(2019)Guo, Lincoln, Davidson, Busch, Yu, Whalen, Harvey, Orts-Escolano, Pandey, Dourgarian, et~al.]{guo2019relightables}
Kaiwen Guo, Peter Lincoln, Philip Davidson, Jay Busch, Xueming Yu, Matt Whalen, Geoff Harvey, Sergio Orts-Escolano, Rohit Pandey, Jason Dourgarian, et~al.
\newblock The relightables: Volumetric performance capture of humans with realistic relighting.
\newblock \emph{ACM Transactions on Graphics (ToG)}, 38\penalty0 (6):\penalty0 1--19, 2019.

\bibitem[Guo et~al.(2024)Guo, Zheng, Hou, Gao, Deng, Wan, Zhang, Liu, Hu, Zha, et~al.]{guo2024i2v}
Xun Guo, Mingwu Zheng, Liang Hou, Yuan Gao, Yufan Deng, Pengfei Wan, Di Zhang, Yufan Liu, Weiming Hu, Zhengjun Zha, et~al.
\newblock I2v-adapter: A general image-to-video adapter for diffusion models.
\newblock In \emph{ACM SIGGRAPH 2024 Conference Papers}, pages 1--12, 2024.

\bibitem[He et~al.(2024{\natexlab{a}})He, Xu, Guo, Wetzstein, Dai, Li, and Yang]{he2024cameractrl}
Hao He, Yinghao Xu, Yuwei Guo, Gordon Wetzstein, Bo Dai, Hongsheng Li, and Ceyuan Yang.
\newblock Cameractrl: Enabling camera control for text-to-video generation.
\newblock \emph{arXiv preprint arXiv:2404.02101}, 2024{\natexlab{a}}.

\bibitem[He et~al.(2024{\natexlab{b}})He, Zhuang, Wang, Yao, Zhu, Li, Zhang, Cao, and Zhu]{he2024head360}
Yuxiao He, Yiyu Zhuang, Yanwen Wang, Yao Yao, Siyu Zhu, Xiaoyu Li, Qi Zhang, Xun Cao, and Hao Zhu.
\newblock Head360: Learning a parametric 3d full-head for free-view synthesis in 360$\circ$.
\newblock In \emph{European Conference on Computer Vision}, pages 254--272. Springer, 2024{\natexlab{b}}.

\bibitem[Hessel et~al.(2021)Hessel, Holtzman, Forbes, Le~Bras, and Choi]{hessel2021clipscore}
Jack Hessel, Ari Holtzman, Maxwell Forbes, Ronan Le~Bras, and Yejin Choi.
\newblock Clipscore: A reference-free evaluation metric for image captioning.
\newblock In \emph{Proceedings of the 2021 Conference on Empirical Methods in Natural Language Processing}, pages 7514--7528, 2021.

\bibitem[Ho and Salimans()]{ho2021classifier}
Jonathan Ho and Tim Salimans.
\newblock Classifier-free diffusion guidance.
\newblock In \emph{NeurIPS 2021 Workshop on Deep Generative Models and Downstream Applications}.

\bibitem[Ho et~al.(2020)Ho, Jain, and Abbeel]{ho2020ddpm}
Jonathan Ho, Ajay Jain, and Pieter Abbeel.
\newblock Denoising diffusion probabilistic models.
\newblock In \emph{Advances in Neural Information Processing Systems}, 2020.

\bibitem[Hong et~al.(2022{\natexlab{a}})Hong, Chen, Lan, Pan, and Liu]{hong2022eva3d}
Fangzhou Hong, Zhaoxi Chen, Yushi Lan, Liang Pan, and Ziwei Liu.
\newblock Eva3d: Compositional 3d human generation from 2d image collections.
\newblock \emph{arXiv preprint arXiv:2210.04888}, 2022{\natexlab{a}}.

\bibitem[Hong et~al.(2022{\natexlab{b}})Hong, Zhang, Pan, Cai, Yang, and Liu]{hong2022avatarclip}
Fangzhou Hong, Mingyuan Zhang, Liang Pan, Zhongang Cai, Lei Yang, and Ziwei Liu.
\newblock Avatarclip: Zero-shot text-driven generation and animation of 3d avatars.
\newblock \emph{ACM Transactions on Graphics}, 2022{\natexlab{b}}.

\bibitem[Hou et~al.(2024)Hou, Wei, Zeng, and Chen]{hou2024training}
Chen Hou, Guoqiang Wei, Yan Zeng, and Zhibo Chen.
\newblock Training-free camera control for video generation.
\newblock \emph{arXiv preprint arXiv:2406.10126}, 2024.

\bibitem[Hu et~al.(2024)Hu, Hong, and Liu]{hu2024structldm}
Tao Hu, Fangzhou Hong, and Ziwei Liu.
\newblock Structldm: Structured latent diffusion for 3d human generation.
\newblock In \emph{European Conference on Computer Vision}, pages 363--381. Springer, 2024.

\bibitem[Huang et~al.(2024)Huang, Shao, Zhang, Zhang, Feng, Liu, and Wang]{huang2023humannorm}
Xin Huang, Ruizhi Shao, Qi Zhang, Hongwen Zhang, Ying Feng, Yebin Liu, and Qing Wang.
\newblock Humannorm: Learning normal diffusion model for high-quality and realistic 3d human generation.
\newblock In \emph{CVPR}, 2024.

\bibitem[Huang et~al.(2023)Huang, Wang, Zeng, Cao, Qi, Shi, Zha, and Zhang]{huang2023dreamwaltz}
Yukun Huang, Jianan Wang, Ailing Zeng, He Cao, Xianbiao Qi, Yukai Shi, Zheng-Jun Zha, and Lei Zhang.
\newblock Dreamwaltz: Make a scene with complex 3d animatable avatars.
\newblock \emph{Advances in Neural Information Processing Systems}, 36:\penalty0 4566--4584, 2023.

\bibitem[Jain et~al.(2022)Jain, Mildenhall, Barron, Abbeel, and Poole]{jain2022zero}
Ajay Jain, Ben Mildenhall, Jonathan~T Barron, Pieter Abbeel, and Ben Poole.
\newblock Zero-shot text-guided object generation with dream fields.
\newblock In \emph{Proceedings of the IEEE/CVF conference on computer vision and pattern recognition}, pages 867--876, 2022.

\bibitem[Jiang et~al.(2023)Jiang, Wang, Zhang, Chai, He, Chen, and Liao]{jiang2023avatarcraft}
Ruixiang Jiang, Can Wang, Jingbo Zhang, Menglei Chai, Mingming He, Dongdong Chen, and Jing Liao.
\newblock Avatarcraft: Transforming text into neural human avatars with parameterized shape and pose control.
\newblock In \emph{Proceedings of the IEEE/CVF International Conference on Computer Vision}, pages 14371--14382, 2023.

\bibitem[Jin et~al.(2025)Jin, Dai, Luo, Baek, and Cho]{jin2025flovd}
Wonjoon Jin, Qi Dai, Chong Luo, Seung-Hwan Baek, and Sunghyun Cho.
\newblock Flovd: Optical flow meets video diffusion model for enhanced camera-controlled video synthesis.
\newblock In \emph{Proceedings of the Computer Vision and Pattern Recognition Conference}, pages 2040--2049, 2025.

\bibitem[Kerbl et~al.(2023)Kerbl, Kopanas, Leimk{\"u}hler, and Drettakis]{kerbl3Dgaussians}
Bernhard Kerbl, Georgios Kopanas, Thomas Leimk{\"u}hler, and George Drettakis.
\newblock 3d gaussian splatting for real-time radiance field rendering.
\newblock \emph{ACM Transactions on Graphics}, 42\penalty0 (4), 2023.

\bibitem[Kirschstein et~al.(2024)Kirschstein, Giebenhain, Tang, Georgopoulos, and Nie{\ss}ner]{kirschstein2024gghead}
Tobias Kirschstein, Simon Giebenhain, Jiapeng Tang, Markos Georgopoulos, and Matthias Nie{\ss}ner.
\newblock Gghead: Fast and generalizable 3d gaussian heads.
\newblock In \emph{SIGGRAPH Asia 2024 Conference Papers}, pages 1--11, 2024.

\bibitem[Kolotouros et~al.(2023)Kolotouros, Alldieck, Zanfir, Bazavan, Fieraru, and Sminchisescu]{kolotouros2023dreamhuman}
Nikos Kolotouros, Thiemo Alldieck, Andrei Zanfir, Eduard Bazavan, Mihai Fieraru, and Cristian Sminchisescu.
\newblock Dreamhuman: Animatable 3d avatars from text.
\newblock \emph{Advances in neural information processing systems}, 36:\penalty0 10516--10529, 2023.

\bibitem[Lan et~al.(2024)Lan, Hong, Yang, Zhou, Meng, Dai, Pan, and Loy]{lan2024ln3diff}
Yushi Lan, Fangzhou Hong, Shuai Yang, Shangchen Zhou, Xuyi Meng, Bo Dai, Xingang Pan, and Chen~Change Loy.
\newblock Ln3diff: Scalable latent neural fields diffusion for speedy 3d generation.
\newblock In \emph{European Conference on Computer Vision}, pages 112--130. Springer, 2024.

\bibitem[Li et~al.(2022)Li, Li, Xiong, and Hoi]{li2022blip}
Junnan Li, Dongxu Li, Caiming Xiong, and Steven Hoi.
\newblock Blip: Bootstrapping language-image pre-training for unified vision-language understanding and generation.
\newblock In \emph{International conference on machine learning}, pages 12888--12900. PMLR, 2022.

\bibitem[Liao et~al.(2024)Liao, Yi, Xiu, Tang, Huang, Thies, and Black]{liao2024tada}
Tingting Liao, Hongwei Yi, Yuliang Xiu, Jiaxiang Tang, Yangyi Huang, Justus Thies, and Michael~J Black.
\newblock Tada! text to animatable digital avatars.
\newblock In \emph{2024 International Conference on 3D Vision (3DV)}, pages 1508--1519. IEEE, 2024.

\bibitem[Lipman et~al.(2023)Lipman, Chen, Ben-Hamu, Nickel, and Le]{lipman2023flow}
Yaron Lipman, Ricky~TQ Chen, Heli Ben-Hamu, Maximilian Nickel, and Matt Le.
\newblock Flow matching for generative modeling.
\newblock In \emph{11th International Conference on Learning Representations, ICLR 2023}, 2023.

\bibitem[Liu et~al.(2024{\natexlab{a}})Liu, Zhan, Tang, Shan, Zeng, Lin, Liu, and Liu]{liu2024humangaussian}
Xian Liu, Xiaohang Zhan, Jiaxiang Tang, Ying Shan, Gang Zeng, Dahua Lin, Xihui Liu, and Ziwei Liu.
\newblock Humangaussian: Text-driven 3d human generation with gaussian splatting.
\newblock In \emph{Proceedings of the IEEE/CVF Conference on Computer Vision and Pattern Recognition}, pages 6646--6657, 2024{\natexlab{a}}.

\bibitem[Liu et~al.(2024{\natexlab{b}})Liu, Zhang, Li, Yan, Gao, Chen, Yuan, Huang, Sun, Gao, et~al.]{liu2024sora}
Yixin Liu, Kai Zhang, Yuan Li, Zhiling Yan, Chujie Gao, Ruoxi Chen, Zhengqing Yuan, Yue Huang, Hanchi Sun, Jianfeng Gao, et~al.
\newblock Sora: A review on background, technology, limitations, and opportunities of large vision models.
\newblock \emph{arXiv preprint arXiv:2402.17177}, 2024{\natexlab{b}}.

\bibitem[Liu et~al.(2022)Liu, Wang, Qi, and Fu]{liu2022towards}
Zhengzhe Liu, Yi Wang, Xiaojuan Qi, and Chi-Wing Fu.
\newblock Towards implicit text-guided 3d shape generation.
\newblock In \emph{Proceedings of the IEEE/CVF Conference on Computer Vision and Pattern Recognition}, pages 17896--17906, 2022.

\bibitem[Loper et~al.(2015)Loper, Mahmood, Romero, Pons-Moll, and Black]{SMPL:2015}
Matthew Loper, Naureen Mahmood, Javier Romero, Gerard Pons-Moll, and Michael~J. Black.
\newblock {SMPL}: A skinned multi-person linear model.
\newblock \emph{ACM Transactions on Graphics}, 2015.

\bibitem[Ma et~al.(2024{\natexlab{a}})Ma, Wang, Jia, Chen, Liu, Li, Chen, and Qiao]{ma2024latte}
Xin Ma, Yaohui Wang, Gengyun Jia, Xinyuan Chen, Ziwei Liu, Yuan-Fang Li, Cunjian Chen, and Yu Qiao.
\newblock Latte: Latent diffusion transformer for video generation.
\newblock \emph{arXiv preprint arXiv:2401.03048}, 2024{\natexlab{a}}.

\bibitem[Ma et~al.(2024{\natexlab{b}})Ma, Lin, Ji, Fan, Sun, and Ji]{ma2024x}
Yiwei Ma, Zhekai Lin, Jiayi Ji, Yijun Fan, Xiaoshuai Sun, and Rongrong Ji.
\newblock X-oscar: a progressive framework for high-quality text-guided 3d animatable avatar generation.
\newblock In \emph{Proceedings of the 41st International Conference on Machine Learning}, pages 33826--33838, 2024{\natexlab{b}}.

\bibitem[Metzer et~al.(2023)Metzer, Richardson, Patashnik, Giryes, and Cohen-Or]{metzer2023latent}
Gal Metzer, Elad Richardson, Or Patashnik, Raja Giryes, and Daniel Cohen-Or.
\newblock Latent-nerf for shape-guided generation of 3d shapes and textures.
\newblock In \emph{Proceedings of the IEEE/CVF Conference on Computer Vision and Pattern Recognition}, pages 12663--12673, 2023.

\bibitem[Michel et~al.(2022)Michel, Bar-On, Liu, Benaim, and Hanocka]{michel2022text2mesh}
Oscar Michel, Roi Bar-On, Richard Liu, Sagie Benaim, and Rana Hanocka.
\newblock Text2mesh: Text-driven neural stylization for meshes.
\newblock In \emph{Proceedings of the IEEE/CVF Conference on Computer Vision and Pattern Recognition}, pages 13492--13502, 2022.

\bibitem[Mohammad~Khalid et~al.(2022)Mohammad~Khalid, Xie, Belilovsky, and Popa]{mohammad2022clip}
Nasir Mohammad~Khalid, Tianhao Xie, Eugene Belilovsky, and Tiberiu Popa.
\newblock Clip-mesh: Generating textured meshes from text using pretrained image-text models.
\newblock In \emph{SIGGRAPH Asia 2022 conference papers}, pages 1--8, 2022.

\bibitem[Pavlakos et~al.(2019)Pavlakos, Choutas, Ghorbani, Bolkart, Osman, Tzionas, and Black]{pavlakos2019expressive}
Georgios Pavlakos, Vasileios Choutas, Nima Ghorbani, Timo Bolkart, Ahmed~AA Osman, Dimitrios Tzionas, and Michael~J Black.
\newblock Expressive body capture: 3d hands, face, and body from a single image.
\newblock In \emph{Proceedings of the IEEE/CVF conference on computer vision and pattern recognition}, pages 10975--10985, 2019.

\bibitem[Poole et~al.(2023)Poole, Jain, Barron, and Mildenhall]{poole2022dreamfusion}
Ben Poole, Ajay Jain, Jonathan~T. Barron, and Ben Mildenhall.
\newblock Dreamfusion: Text-to-3d using 2d diffusion.
\newblock In \emph{ICLR}, 2023.

\bibitem[Radford et~al.(2021)Radford, Kim, Hallacy, Ramesh, Goh, Agarwal, Sastry, Askell, Mishkin, Clark, et~al.]{radford2021learning}
Alec Radford, Jong~Wook Kim, Chris Hallacy, Aditya Ramesh, Gabriel Goh, Sandhini Agarwal, Girish Sastry, Amanda Askell, Pamela Mishkin, Jack Clark, et~al.
\newblock Learning transferable visual models from natural language supervision.
\newblock In \emph{International conference on machine learning}, pages 8748--8763. PmLR, 2021.

\bibitem[Ramesh et~al.(2022)Ramesh, Dhariwal, Nichol, Chu, and Chen]{ramesh2022dalle2}
Aditya Ramesh, Prafulla Dhariwal, Alex Nichol, Casey Chu, and Mark Chen.
\newblock Hierarchical text-conditional image generation with clip latents.
\newblock \emph{arXiv preprint arXiv:2204.06125}, 2022.

\bibitem[Rombach et~al.(2022{\natexlab{a}})Rombach, Blattmann, Lorenz, Esser, and Ommer]{latentdiffusion}
Robin Rombach, Andreas Blattmann, Dominik Lorenz, Patrick Esser, and Bj{\"o}rn Ommer.
\newblock High-resolution image synthesis with latent diffusion models.
\newblock In \emph{Proceedings of the IEEE/CVF conference on computer vision and pattern recognition}, pages 10684--10695, 2022{\natexlab{a}}.

\bibitem[Rombach et~al.(2022{\natexlab{b}})Rombach, Blattmann, Lorenz, Esser, and Ommer]{rombach2022high}
Robin Rombach, Andreas Blattmann, Dominik Lorenz, Patrick Esser, and Bj{\"o}rn Ommer.
\newblock High-resolution image synthesis with latent diffusion models.
\newblock In \emph{Proceedings of the IEEE/CVF conference on computer vision and pattern recognition}, pages 10684--10695, 2022{\natexlab{b}}.

\bibitem[Rombach et~al.(2022{\natexlab{c}})Rombach, Blattmann, Lorenz, Esser, and Ommer]{rombach2022ldm}
Robin Rombach, Andreas Blattmann, Dominik Lorenz, Patrick Esser, and Bj{\"o}rn Ommer.
\newblock High-resolution image synthesis with latent diffusion models.
\newblock In \emph{IEEE Conference on Computer Vision and Pattern Recognition}, 2022{\natexlab{c}}.

\bibitem[Saharia et~al.(2022)Saharia, Chan, Saxena, Li, Whang, Denton, Ghasemipour, Gontijo~Lopes, Karagol~Ayan, Salimans, et~al.]{saharia2022photorealistic}
Chitwan Saharia, William Chan, Saurabh Saxena, Lala Li, Jay Whang, Emily~L Denton, Kamyar Ghasemipour, Raphael Gontijo~Lopes, Burcu Karagol~Ayan, Tim Salimans, et~al.
\newblock Photorealistic text-to-image diffusion models with deep language understanding.
\newblock \emph{Advances in neural information processing systems}, 35:\penalty0 36479--36494, 2022.

\bibitem[Saito et~al.(2019)Saito, Huang, Natsume, Morishima, Kanazawa, and Li]{saito2019pifu}
Shunsuke Saito, Zeng Huang, Ryota Natsume, Shigeo Morishima, Angjoo Kanazawa, and Hao Li.
\newblock Pifu: Pixel-aligned implicit function for high-resolution clothed human digitization.
\newblock In \emph{Proceedings of the IEEE/CVF international conference on computer vision}, pages 2304--2314, 2019.

\bibitem[Shu et~al.(2019)Shu, Park, and Kwon]{shu20193d}
Dong~Wook Shu, Sung~Woo Park, and Junseok Kwon.
\newblock 3d point cloud generative adversarial network based on tree structured graph convolutions.
\newblock In \emph{Proceedings of the IEEE/CVF international conference on computer vision}, pages 3859--3868, 2019.

\bibitem[Tang et~al.(2024)Tang, Ren, Zhou, Liu, and Zeng]{tang2023dreamgaussian}
Jiaxiang Tang, Jiawei Ren, Hang Zhou, Ziwei Liu, and Gang Zeng.
\newblock Dreamgaussian: Generative gaussian splatting for efficient 3d content creation.
\newblock In \emph{ICLR}, 2024.

\bibitem[Tian et~al.(2023)Tian, Yang, and Wu]{tian2023shapescaffolder}
Xi Tian, Yong-Liang Yang, and Qi Wu.
\newblock Shapescaffolder: Structure-aware 3d shape generation from text.
\newblock In \emph{Proceedings of the IEEE/CVF International Conference on Computer Vision}, pages 2715--2724, 2023.

\bibitem[Tochilkin et~al.(2024)Tochilkin, Pankratz, Liu, Huang, , Letts, Li, Liang, Laforte, Jampani, and Cao]{TripoSR2024}
Dmitry Tochilkin, David Pankratz, Zexiang Liu, Zixuan Huang, , Adam Letts, Yangguang Li, Ding Liang, Christian Laforte, Varun Jampani, and Yan-Pei Cao.
\newblock Triposr: Fast 3d object reconstruction from a single image.
\newblock \emph{arXiv preprint arXiv:2403.02151}, 2024.

\bibitem[Wang et~al.(2022)Wang, Chai, He, Chen, and Liao]{wang2022clip}
Can Wang, Menglei Chai, Mingming He, Dongdong Chen, and Jing Liao.
\newblock Clip-nerf: Text-and-image driven manipulation of neural radiance fields.
\newblock In \emph{Proceedings of the IEEE/CVF Conference on Computer Vision and Pattern Recognition}, pages 3835--3844, 2022.

\bibitem[Wang et~al.(2023{\natexlab{a}})Wang, Du, Li, Yeh, and Shakhnarovich]{wang2023score}
Haochen Wang, Xiaodan Du, Jiahao Li, Raymond~A Yeh, and Greg Shakhnarovich.
\newblock Score jacobian chaining: Lifting pretrained 2d diffusion models for 3d generation.
\newblock In \emph{Proceedings of the IEEE/CVF Conference on Computer Vision and Pattern Recognition}, pages 12619--12629, 2023{\natexlab{a}}.

\bibitem[Wang et~al.(2024)Wang, Liu, Dou, Yu, Liang, Lin, Xie, Song, Li, and Wang]{wang2024disentangled}
Jionghao Wang, Yuan Liu, Zhiyang Dou, Zhengming Yu, Yongqing Liang, Cheng Lin, Rong Xie, Li Song, Xin Li, and Wenping Wang.
\newblock Disentangled clothed avatar generation from text descriptions.
\newblock In \emph{European Conference on Computer Vision}, pages 381--401, 2024.

\bibitem[Wang et~al.(2023{\natexlab{b}})Wang, Lu, Wang, Bao, Li, Su, and Zhu]{wang2023prolificdreamer}
Zhengyi Wang, Cheng Lu, Yikai Wang, Fan Bao, Chongxuan Li, Hang Su, and Jun Zhu.
\newblock Prolificdreamer: High-fidelity and diverse text-to-3d generation with variational score distillation.
\newblock \emph{Advances in neural information processing systems}, 36:\penalty0 8406--8441, 2023{\natexlab{b}}.

\bibitem[Wei et~al.(2023)Wei, Wang, Feng, Lin, and Yap]{wei2023taps3d}
Jiacheng Wei, Hao Wang, Jiashi Feng, Guosheng Lin, and Kim-Hui Yap.
\newblock Taps3d: Text-guided 3d textured shape generation from pseudo supervision.
\newblock In \emph{Proceedings of the IEEE/CVF conference on computer vision and pattern recognition}, pages 16805--16815, 2023.

\bibitem[Wu et~al.(2016)Wu, Zhang, Xue, Freeman, and Tenenbaum]{wu2016learning}
Jiajun Wu, Chengkai Zhang, Tianfan Xue, Bill Freeman, and Josh Tenenbaum.
\newblock Learning a probabilistic latent space of object shapes via 3d generative-adversarial modeling.
\newblock \emph{Advances in neural information processing systems}, 29, 2016.

\bibitem[Wu et~al.(2024)Wu, Lin, Zhang, Zeng, Xu, Torr, Cao, and Yao]{direct3d}
Shuang Wu, Youtian Lin, Feihu Zhang, Yifei Zeng, Jingxi Xu, Philip Torr, Xun Cao, and Yao Yao.
\newblock Direct3d: Scalable image-to-3d generation via 3d latent diffusion transformer.
\newblock \emph{Advances in Neural Information Processing Systems}, 37:\penalty0 121859--121881, 2024.

\bibitem[Xiang et~al.(2025)Xiang, Lv, Xu, Deng, Wang, Zhang, Chen, Tong, and Yang]{xiang2025structured}
Jianfeng Xiang, Zelong Lv, Sicheng Xu, Yu Deng, Ruicheng Wang, Bowen Zhang, Dong Chen, Xin Tong, and Jiaolong Yang.
\newblock Structured 3d latents for scalable and versatile 3d generation.
\newblock In \emph{Proceedings of the Computer Vision and Pattern Recognition Conference}, pages 21469--21480, 2025.

\bibitem[Xiu et~al.(2023)Xiu, Yang, Cao, Tzionas, and Black]{xiu2023econ}
Yuliang Xiu, Jinlong Yang, Xu Cao, Dimitrios Tzionas, and Michael~J Black.
\newblock Econ: Explicit clothed humans optimized via normal integration.
\newblock In \emph{Proceedings of the IEEE/CVF conference on computer vision and pattern recognition}, pages 512--523, 2023.

\bibitem[Yang et~al.(2024)Yang, Yang, Zhang, Hui, Zheng, Yu, Li, Liu, Huang, Wei, et~al.]{yang2024qwen2}
An Yang, Baosong Yang, Beichen Zhang, Binyuan Hui, Bo Zheng, Bowen Yu, Chengyuan Li, Dayiheng Liu, Fei Huang, Haoran Wei, et~al.
\newblock Qwen2. 5 technical report.
\newblock \emph{arXiv preprint arXiv:2412.15115}, 2024.

\bibitem[Yi et~al.(2023)Yi, Fang, Wu, Xie, Zhang, Liu, Tian, and Wang]{yi2023gaussiandreamer}
Taoran Yi, Jiemin Fang, Guanjun Wu, Lingxi Xie, Xiaopeng Zhang, Wenyu Liu, Qi Tian, and Xinggang Wang.
\newblock Gaussiandreamer: Fast generation from text to 3d gaussian splatting with point cloud priors.
\newblock \emph{arXiv preprint arXiv:2310.08529}, 2023.

\bibitem[Yu et~al.(2018)Yu, Zheng, Guo, Zhao, Dai, Li, Pons-Moll, and Liu]{yu2018doublefusion}
Tao Yu, Zerong Zheng, Kaiwen Guo, Jianhui Zhao, Qionghai Dai, Hao Li, Gerard Pons-Moll, and Yebin Liu.
\newblock Doublefusion: Real-time capture of human performances with inner body shapes from a single depth sensor.
\newblock In \emph{Proceedings of the IEEE conference on computer vision and pattern recognition}, pages 7287--7296, 2018.

\bibitem[Yu et~al.(2021)Yu, Zheng, Guo, Liu, Dai, and Liu]{yu2021function4d}
Tao Yu, Zerong Zheng, Kaiwen Guo, Pengpeng Liu, Qionghai Dai, and Yebin Liu.
\newblock Function4d: Real-time human volumetric capture from very sparse consumer rgbd sensors.
\newblock In \emph{Proceedings of the IEEE/CVF conference on computer vision and pattern recognition}, pages 5746--5756, 2021.

\bibitem[Zeng et~al.(2023)Zeng, Lu, Ji, Yao, Zhu, and Cao]{zeng2023avatarbooth}
Yifei Zeng, Yuanxun Lu, Xinya Ji, Yao Yao, Hao Zhu, and Xun Cao.
\newblock Avatarbooth: High-quality and customizable 3d human avatar generation.
\newblock \emph{arXiv preprint arXiv:2306.09864}, 2023.

\bibitem[Zhang et~al.(2023{\natexlab{a}})Zhang, Feng, Kulits, Wen, Thies, and Black]{zhang2023text}
Hao Zhang, Yao Feng, Peter Kulits, Yandong Wen, Justus Thies, and Michael~J Black.
\newblock Text-guided generation and editing of compositional 3d avatars.
\newblock \emph{arXiv preprint arXiv:2309.07125}, 2023{\natexlab{a}}.

\bibitem[Zhang et~al.(2024{\natexlab{a}})Zhang, Chen, Yang, Qu, Wang, Chen, Long, Zhu, Du, and Zheng]{zhang2023avatarverse}
Huichao Zhang, Bowen Chen, Hao Yang, Liao Qu, Xu Wang, Li Chen, Chao Long, Feida Zhu, Daniel Du, and Min Zheng.
\newblock Avatarverse: High-quality \& stable 3d avatar creation from text and pose.
\newblock In \emph{Proceedings of the AAAI Conference on Artificial Intelligence}, pages 7124--7132, 2024{\natexlab{a}}.

\bibitem[Zhang et~al.(2025)Zhang, Wu, Liang, Gong, Hu, Yao, Cao, and Zhu]{zhang2025fate}
Jiawei Zhang, Zijian Wu, Zhiyang Liang, Yicheng Gong, Dongfang Hu, Yao Yao, Xun Cao, and Hao Zhu.
\newblock Fate: Full-head gaussian avatar with textural editing from monocular video.
\newblock 2025.

\bibitem[Zhang et~al.(2023{\natexlab{b}})Zhang, Rao, and Agrawala]{zhang2023adding}
Lvmin Zhang, Anyi Rao, and Maneesh Agrawala.
\newblock Adding conditional control to text-to-image diffusion models.
\newblock In \emph{Proceedings of the IEEE/CVF international conference on computer vision}, pages 3836--3847, 2023{\natexlab{b}}.

\bibitem[Zhang et~al.(2024{\natexlab{b}})Zhang, Wang, Zhang, Qiu, Pang, Jiang, Yang, Xu, and Yu]{clay}
Longwen Zhang, Ziyu Wang, Qixuan Zhang, Qiwei Qiu, Anqi Pang, Haoran Jiang, Wei Yang, Lan Xu, and Jingyi Yu.
\newblock Clay: A controllable large-scale generative model for creating high-quality 3d assets.
\newblock \emph{ACM Transactions on Graphics}, 2024{\natexlab{b}}.

\bibitem[Zhang et~al.(2024{\natexlab{c}})Zhang, Yan, Liu, Sheng, and Yang]{zhang20243gen}
Weitian Zhang, Yichao Yan, Yunhui Liu, Xingdong Sheng, and Xiaokang Yang.
\newblock E3gen: Efficient, expressive and editable avatars generation.
\newblock In \emph{Proceedings of the 32nd ACM International Conference on Multimedia}, pages 6860--6869, 2024{\natexlab{c}}.

\bibitem[Zhao et~al.(2023{\natexlab{a}})Zhao, Li, Hu, Li, Zou, Shi, and Fan]{zhao2023zero}
Rui Zhao, Wei Li, Zhipeng Hu, Lincheng Li, Zhengxia Zou, Zhenwei Shi, and Changjie Fan.
\newblock Zero-shot text-to-parameter translation for game character auto-creation.
\newblock In \emph{Proceedings of the IEEE/CVF Conference on Computer Vision and Pattern Recognition}, pages 21013--21023, 2023{\natexlab{a}}.

\bibitem[Zhao et~al.(2023{\natexlab{b}})Zhao, Chen, Chen, Bao, Hao, Yuan, and Wong]{zhao2023uni}
Shihao Zhao, Dongdong Chen, Yen-Chun Chen, Jianmin Bao, Shaozhe Hao, Lu Yuan, and Kwan-Yee~K Wong.
\newblock Uni-controlnet: All-in-one control to text-to-image diffusion models.
\newblock \emph{Advances in Neural Information Processing Systems}, 36:\penalty0 11127--11150, 2023{\natexlab{b}}.

\bibitem[Zheng et~al.(2024)Zheng, Li, Jiang, Lu, Wu, and Li]{zheng2024cami2v}
Guangcong Zheng, Teng Li, Rui Jiang, Yehao Lu, Tao Wu, and Xi Li.
\newblock Cami2v: Camera-controlled image-to-video diffusion model.
\newblock \emph{arXiv preprint arXiv:2410.15957}, 2024.

\bibitem[Zheng et~al.(2021)Zheng, Yu, Liu, and Dai]{zheng2021pamir}
Zerong Zheng, Tao Yu, Yebin Liu, and Qionghai Dai.
\newblock Pamir: Parametric model-conditioned implicit representation for image-based human reconstruction.
\newblock \emph{IEEE transactions on pattern analysis and machine intelligence}, 44\penalty0 (6):\penalty0 3170--3184, 2021.

\bibitem[Zhu et~al.(2021)Zhu, Zuo, Yang, Wang, Cao, and Yang]{zhu2021detailed}
Hao Zhu, Xinxin Zuo, Haotian Yang, Sen Wang, Xun Cao, and Ruigang Yang.
\newblock Detailed avatar recovery from single image.
\newblock \emph{IEEE Transactions on Pattern Analysis and Machine Intelligence}, 44\penalty0 (11):\penalty0 7363--7379, 2021.

\bibitem[Zhu et~al.(2024)Zhu, Chen, Dai, Dong, Xu, Cao, Yao, Zhu, and Zhu]{zhu2024champ}
Shenhao Zhu, Junming~Leo Chen, Zuozhuo Dai, Zilong Dong, Yinghui Xu, Xun Cao, Yao Yao, Hao Zhu, and Siyu Zhu.
\newblock Champ: Controllable and consistent human image animation with 3d parametric guidance.
\newblock In \emph{European Conference on Computer Vision}, pages 145--162. Springer, 2024.

\bibitem[Zhuang et~al.(2025{\natexlab{a}})Zhuang, Kang, Bao, Lin, and Li]{zhuang2024dagsm}
Jingyu Zhuang, Di Kang, Linchao Bao, Liang Lin, and Guanbin Li.
\newblock Dagsm: Disentangled avatar generation with gs-enhanced mesh.
\newblock In \emph{Proceedings of the Computer Vision and Pattern Recognition Conference}, pages 292--303, 2025{\natexlab{a}}.

\bibitem[Zhuang et~al.(2022)Zhuang, Zhu, Sun, and Cao]{zhuang2022mofanerf}
Yiyu Zhuang, Hao Zhu, Xusen Sun, and Xun Cao.
\newblock Mofanerf: Morphable facial neural radiance field.
\newblock In \emph{European conference on computer vision}, pages 268--285. Springer, 2022.

\bibitem[Zhuang et~al.(2024)Zhuang, He, Zhang, Wang, Zhu, Yao, Zhu, Cao, and Zhu]{zhuang2024towards}
Yiyu Zhuang, Yuxiao He, Jiawei Zhang, Yanwen Wang, Jiahe Zhu, Yao Yao, Siyu Zhu, Xun Cao, and Hao Zhu.
\newblock Towards native generative model for 3d head avatar.
\newblock \emph{arXiv preprint arXiv:2410.01226}, 2024.

\bibitem[Zhuang et~al.(2025{\natexlab{b}})Zhuang, Lv, Wen, Shuai, Zeng, Zhu, Chen, Yang, Cao, and Liu]{zhuang2025idol}
Yiyu Zhuang, Jiaxi Lv, Hao Wen, Qing Shuai, Ailing Zeng, Hao Zhu, Shifeng Chen, Yujiu Yang, Xun Cao, and Wei Liu.
\newblock Idol: Instant photorealistic 3d human creation from a single image.
\newblock In \emph{Proceedings of the IEEE/CVF Conference on Computer Vision and Pattern Recognition}, 2025{\natexlab{b}}.

\end{thebibliography}
